\documentclass[sigconf,nonacm]{acmart}

\usepackage{booktabs}
\usepackage{array}
\usepackage{makecell}
\usepackage[table]{xcolor}
\usepackage{bbding}
\usepackage{graphicx}
\usepackage{multirow}
\usepackage{enumitem}
\usepackage{soul}
\setul{1.5pt}{0.4pt}

\begin{document}

\title{ClinVerse: A Realistic Multi-Agent Diagnostic Environment for Training and Evaluating Medical Agents}

\author{Sirry Chen}
\orcid{0009-0004-0568-2677}
\affiliation{%
  \institution{Fudan University}
  \city{Shanghai}
  \country{China}
}
\affiliation{%
  \institution{Shanghai Innovation Institute}
  \city{Shanghai}
  \country{China}
}
\email{siyuanchen25@m.fudan.edu.cn}

\author{Fangke Chen}
\affiliation{%
  \institution{Zhejiang University}
  \city{Hangzhou}
  \country{China}
}
\affiliation{%
  \institution{Shanghai Innovation Institute}
  \city{Shanghai}
  \country{China}
}
\email{fkchen@zju.edu.cn}

\author{Xuanwen Ding}
\affiliation{%
  \institution{Fudan University}
  \city{Shanghai}
  \country{China}
}
\affiliation{%
  \institution{Shanghai Innovation Institute}
  \city{Shanghai}
  \country{China}
}
\email{xwding25@m.fudan.edu.cn}

\author{Wei Chen}
\affiliation{%
 \institution{Huazhong University of Science and Technology}
 \city{Wuhan}
 \country{China}}
\email{lemuria\_chen@hust.edu.cn}

\author{Zhongyu Wei}
\correspondingauthor
\affiliation{%
  \institution{School of Data Science, \\Fudan University}
  \city{Shanghai}
  \country{China}
}
\affiliation{%
  \institution{Shanghai Innovation Institute}
  \city{Shanghai}
  \country{China}
}
\email{zywei@fudan.edu.cn}

\renewcommand{\shortauthors}{Chen et al.}

\begin{abstract}

Disease diagnosis is a key application of large language model (LLM) agents in healthcare. However, existing medical agents are primarily trained and evaluated with static question-answering data, providing limited opportunities to learn active information acquisition, examination decision-making, and dynamic clinical interaction.
To address this limitation, we present \textbf{\textsc{ClinVerse}}, an interactive, real-data-grounded multi-agent environment designed for training and evaluating medical diagnostic agents. Built on \textbf{\textsc{ClinPatient}}, a scalable dataset comprising 150k diagnostic cases derived from real-world electronic health records, \textsc{ClinVerse} supports coordinated interactions among six specialized clinical roles.
Specifically, a doctor agent starts from an initial patient complaint, progressively gathers clinical evidence through patient interviews and examination requests, and ultimately produces a diagnosis. For each interaction trajectory, a rule-based scoring framework quantifies information acquisition, examination ordering, and diagnostic accuracy. These trajectory-level scores serve as reward signals for agent training while also enabling scalable evaluation.
In experiments, we evaluate various general-purpose and medical-domain models and show that nine self-evolution methods across three categories consistently improve agents' diagnostic capabilities within \textsc{ClinVerse}. We further analyze the interplay among model capability, environment feedback, and self-evolution methods\footnote{Source code is available at \href{https://github.com/SirryChen/ClinVerse}{\textcolor{blue}{\textit{github}}}, demo at \href{https://sirrychen.github.io/blogs/ClinVerse/}{\textcolor{blue}{\textit{website}}}.}.

\end{abstract}

\settopmatter{printacmref=false, printccs=false, printfolios=true}
\maketitle

\section{Introduction}

\newcommand{\yes}{\textcolor{green!60!black}{\CheckmarkBold}}
\newcommand{\no}{\textcolor{red!80!black}{\XSolidBrush}}

\newcolumntype{L}[1]{>{\raggedright\arraybackslash}p{#1}}
\newcolumntype{C}[1]{>{\centering\arraybackslash}p{#1}}

\DeclareRobustCommand{\roleicon}[1]{%
  \raisebox{-0.28\height}{%
    \includegraphics[height=2.8ex]{figure/#1.png}%
  }%
}

\newcommand{\doctoricon}{\roleicon{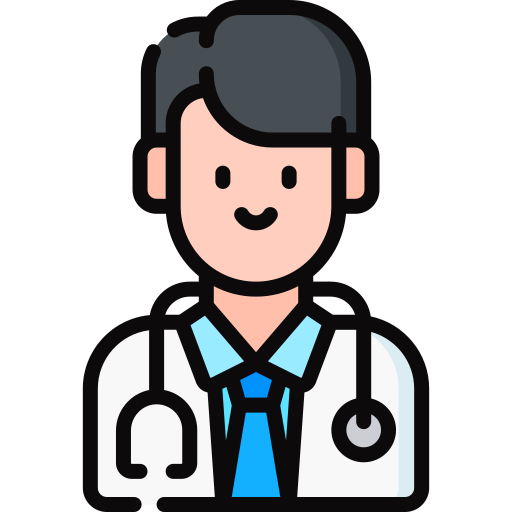}}
\newcommand{\judgeicon}{\roleicon{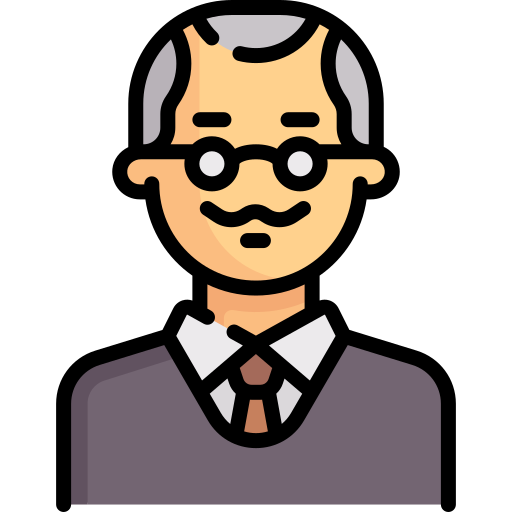}}
\newcommand{\labicon}{\roleicon{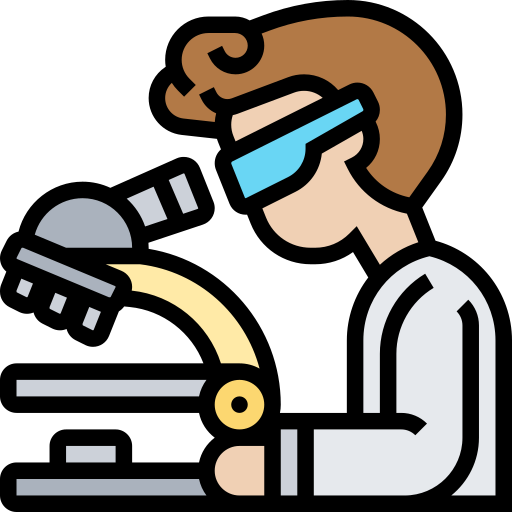}}
\newcommand{\nurseicon}{\roleicon{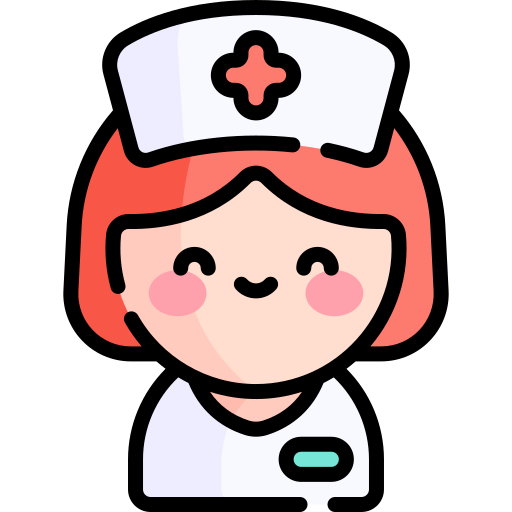}}
\newcommand{\patienticon}{\roleicon{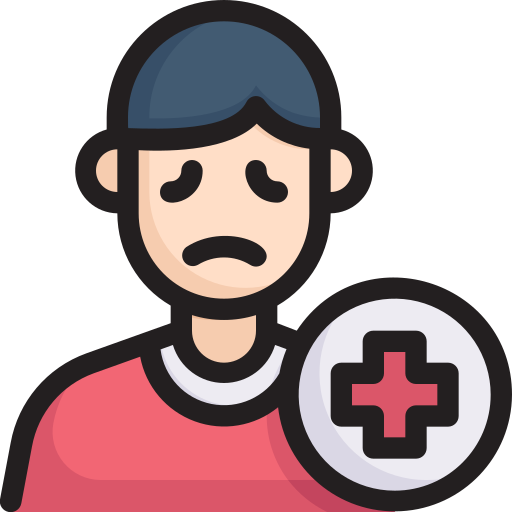}}
\newcommand{\controllericon}{\roleicon{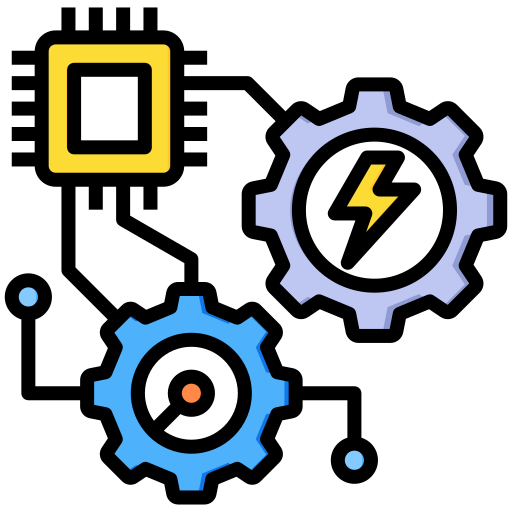}}
\newcommand{\latentpatienticon}{\roleicon{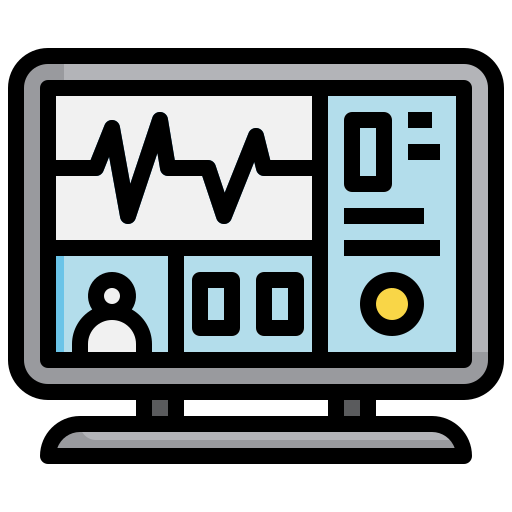}}

\newcommand{\iconsep}{\hspace{0.30em}}

\begin{table*}[t]
  \caption[
    Comparison of static medical QA and interactive clinical
    environments.
  ]{
    Comparison of static medical QA and
    interactive clinical environments. \textsc{ClinVerse} is grounded in 150k scalable real-world patient data, and supports predicting missing examination results to enable flexible interactions of 6 roles.
    Role icons denote
    \protect\doctoricon{} medical agent,
    \protect\labicon{} laboratory physician,
    \protect\nurseicon{} nurse,
    \protect\patienticon{} \protect\latentpatienticon{} patient,
    \protect\judgeicon{} judge, and
    \protect\controllericon{} environment controller. SME means supporting missing examinations.
  }
  \label{tab:benchmark_comparison}

  \centering
  \small
  \setlength{\tabcolsep}{3.2pt}
  \renewcommand{\arraystretch}{1.15}

  \begin{tabular}{
    @{}
    L{2.3cm}
    l
    L{0.90cm}
    L{3.7cm}
    C{2.7cm}
    C{1.5cm}
    C{1.5cm}
    C{1.5cm}
    @{}
  }
    \toprule

    \textbf{Work}
    & \textbf{Venue}
    & \textbf{Cases}
    & \textbf{Data Source}
    & \textbf{Roles}
    & \textbf{Scalable}
    & \makecell[c]{\textbf{Real Data}}
    & \makecell[c]{\textbf{SME}} \\

    \midrule

    \rowcolor{gray!15}
    \multicolumn{8}{@{}l}{\textbf{Static Question-answering}} \\

    MedQA~\citep{MedQA}
    & arXiv 2020
    & 61k
    & Medical licensing exams
    & ---
    & \no
    & \yes
    & \no \\

    CMExam~\citep{CMExam}
    & NeurIPS 2023
    & 68k
    & Chinese medical licensing exam
    & ---
    & \no
    & \yes
    & \no \\

    MedXpertQA~\citep{MedXpertQA}
    & ICML 2025
    & 4,460
    & Medical examinations
    & ---
    & \no
    & \yes
    & \no \\

    EHRBench~\citep{EHRBench}
    & KDD 2026
    & 960k
    & MIMIC-III, MIMIC-IV, etc.
    & ---
    & \yes
    & \yes
    & \no \\

    \midrule

    \rowcolor{gray!15}
    \multicolumn{8}{@{}l}{\textbf{Interactive Clinical Environments}} \\

    Agent Hospital~\citep{Agent_Hospital}
    & arXiv 2024
    & 10k
    & LLM-generated cases
    & \doctoricon
      \iconsep\nurseicon
      \iconsep\patienticon
      \iconsep\controllericon
    & \yes
    & \no
    & \yes \\

    AIHospital~\citep{AI_Hospital}
    & COLING 2025
    & 506
    & Web-based clinical cases
    & \doctoricon
      \iconsep\patienticon
      \iconsep\controllericon
      \iconsep\judgeicon
    & \no
    & \yes
    & \no \\

    SDBench~\citep{SDBench}
    & Microsoft AI 2025
    & 304
    & NEJM-CPC
    & \doctoricon
      \iconsep\patienticon
      \iconsep\judgeicon
    & \no
    & \yes
    & \yes \\

    MedSP~\citep{MedSP}
    & arXiv 2026
    & 1,638
    & MedEdPORTAL
    & \doctoricon
      \iconsep\patienticon
      \iconsep\controllericon
    & \no
    & \yes
    & \no \\

    MIRA~\citep{MIRA}
    & Nature 2026
    & 574
    & Expert-processed MIMIC-IV
    & \doctoricon
      \iconsep\patienticon
      \iconsep\judgeicon
    & \no
    & \yes
    & \no \\

    \midrule

    \textbf{ClinVerse}
    & \textbf{Ours}
    & \textbf{150k}
    & \textbf{MIMIC-IV, SICdb, etc.}
    & \doctoricon
      \iconsep\labicon
      \iconsep\patienticon
      \iconsep\latentpatienticon
      \iconsep\judgeicon
      \iconsep\controllericon
    & \yes
    & \yes
    & \yes \\

    \bottomrule
  \end{tabular}
\end{table*}

To alleviate the shortage of healthcare resources, medical agents driven by large language models (LLMs) are increasingly being integrated into modern clinical practice, with assisting diagnosis emerging as one of their most important applications~\citep{medical_stages,diag_important}. To ensure their reliability in real-world clinical settings, extensive research has been
devoted to evaluating and advancing their diagnostic capabilities ~\citep{singhal2023clinical,
hager2024evaluation,tu2025conversational}.

As shown in Table~\ref{tab:benchmark_comparison}, early studies predominantly assess and train medical agents within static single-turn question-answering (QA) formats derived from medical licensing exams~\citep{MedQA,CMExam,MedXpertQA,EHRBench}. This paradigm provides medical agents with pre-defined clinical information and requires them to directly formulate a diagnosis, offering standardization and efficiency in model training and evaluation. However, in real-world clinical practice, \ul{physicians must actively gather information through multi-turn patient consultations, coordinate with laboratory and radiology departments to request diagnostic tests, and progressively refine their diagnostic hypotheses based on newly acquired evidence}~\citep{NASEM2015Diagnosis,kassirer1983teaching}. This fundamental discrepancy highlights that static QA tasks primarily assess medical agents' diagnostic reasoning capabilities while overlooking their information acquisition capabilities, thereby failing to fully capture the dynamic and interactive nature of clinical diagnosis.

These limitations motivate a shift from static QA toward interactive simulations of real-world clinical environments that better reflect real-world multi-role interactions. Within such environments, medical agents can continuously improve their diagnostic reasoning capability and information acquisition capability, through iterative interaction, feedback, and self-evolution. Compared with static QA, this shifts the training paradigm from \textit{taking medical exams} to \textit{undergoing clinical residency training}.

Despite this promising vision, constructing interactive clinical environments remains challenging.
The first challenge is \textbf{how to prevent information leakage while supporting open-ended clinical interactions}. During patient simulation, granting simulators access to complete clinical records may inadvertently expose ground-truth diagnoses or unrevealed examination results, undermining the authenticity of clinical interactions. Moreover, when agents request examinations absent from electronic health records (EHRs), existing environments~\citep{AI_Hospital,SDBench,MedSP} often return "unavailable" or "normal", potentially exposing missingness patterns that agents can exploit as diagnostic shortcuts. These issues highlight the difficulty of preserving controlled information access while enabling flexible clinical interactions.

The second challenge is \textbf{how to provide reliable trajectory-level feedback for evaluating and improving medical agents}. Unlike static QA tasks that primarily assess final-answer correctness, interactive clinical diagnosis involves multi-turn trajectories of information acquisition, examination decisions, and diagnostic reasoning, requiring evaluation of both the clinical decision-making process and its outcomes~\citep{tu2025conversational,mehandru2024evaluating}. However, expert-based evaluation, as adopted in MIRA~\citep{MIRA}, is costly and difficult to scale, while LLM-as-a-judge approaches~\citep{AI_Hospital,Agent_Hospital} may introduce model-dependent biases and sensitivity to prompting strategies~\citep{zheng2023judging,wang2024fair}.

To address these challenges, we propose \textbf{\textsc{ClinVerse}}, an interactive, EHR-grounded multi-agent environment designed for training and evaluating medical diagnostic agents.
To achieve scalability and open-endedness while preserving clinical grounding, we develop an \textbf{EHR-to-environment transformation framework} that converts real-world clinical records into interactive diagnostic environments. Specifically, an automated data construction pipeline enables the scalable construction of \textbf{\textsc{ClinPatient}}, a dataset comprising 150k EHR-derived patient cases, reducing dependence on extensive manual curation. We further introduce a dual-layer patient simulator with a self-check mechanism that separates patient-observable information from latent clinical states, mitigating unintended disclosure of privileged evidence. Meanwhile, missing examination results in EHRs are addressed by \textbf{\textsc{ClinLab}}, a trained physiological-state predictor that infers clinically plausible results.

To provide scalable and reproducible trajectory-level feedback, we develop a \textbf{rule-based evaluation framework} grounded in structured clinical evidence and ICD-10 diagnosis labels. Rather than relying on subjective judgments of entire conversations, this framework evaluates interaction trajectories using deterministic rules across three complementary dimensions: information acquisition, examination coverage, and diagnostic accuracy. These metrics characterize both the evidence acquisition process and the resulting diagnostic outcomes, providing consistent feedback for agent evaluation and subsequent self-evolution.

Built upon these components, \textsc{ClinVerse} coordinates six specialized roles to simulate the core workflow of clinical diagnosis, where \ul{a doctor agent, coordinated by the environment controller, progressively acquires patient information through multi-turn consultations with the observable patient simulator, requests examination results from the laboratory physician supported by the latent patient simulator, and ultimately formulates a diagnosis. The resulting interaction trajectory is then evaluated by a judge.}

Consequently, \textsc{ClinVerse} can serve as both an evaluation platform and a training environment for medical agents. We evaluate nine general-purpose and medical-domain agents, identifying their strengths and limitations in interactive diagnosis. We further investigate multiple self-evolution methods across three levels, demonstrating that agents can improve their capabilities by learning from interaction trajectories generated within \textsc{ClinVerse}. Beyond performance improvements, our analyses reveal that evidence acquisition remains a major bottleneck, while collecting more information does not necessarily lead to better diagnoses, highlighting the importance of evidence prioritization and integration. We also examine how trajectory effectiveness depends on the capabilities of source and target models, and whether learned skills transfer to other clinical environments.

Our main contributions are summarized as follows:

\begin{itemize}[leftmargin=1.5em, labelsep=0.5em]
    \item We introduce \textbf{\textsc{ClinVerse}}, a realistic diagnostic environment that supports open-ended multi-turn interactions among six specialized roles and enables both the self-evolving and evaluation of medical agents.

    \item We develop a scalable pipeline for transforming EHR data into diagnostic cases and use it to construct \textbf{\textsc{ClinPatient}}, a dataset containing \textbf{150k} patient cases. We design a dual-layer patient simulator with a self-check mechanism and train a physiological-state predictor to infer missing examination results.

    \item We demonstrate the effectiveness of \textsc{ClinVerse} as a training environment and validate its reliability as an evaluation platform, and further analyze the interplay among the environment, model, and self-evolution methods..
\end{itemize}

\begin{figure*}[t]
    \centering
    \includegraphics[width=\textwidth]{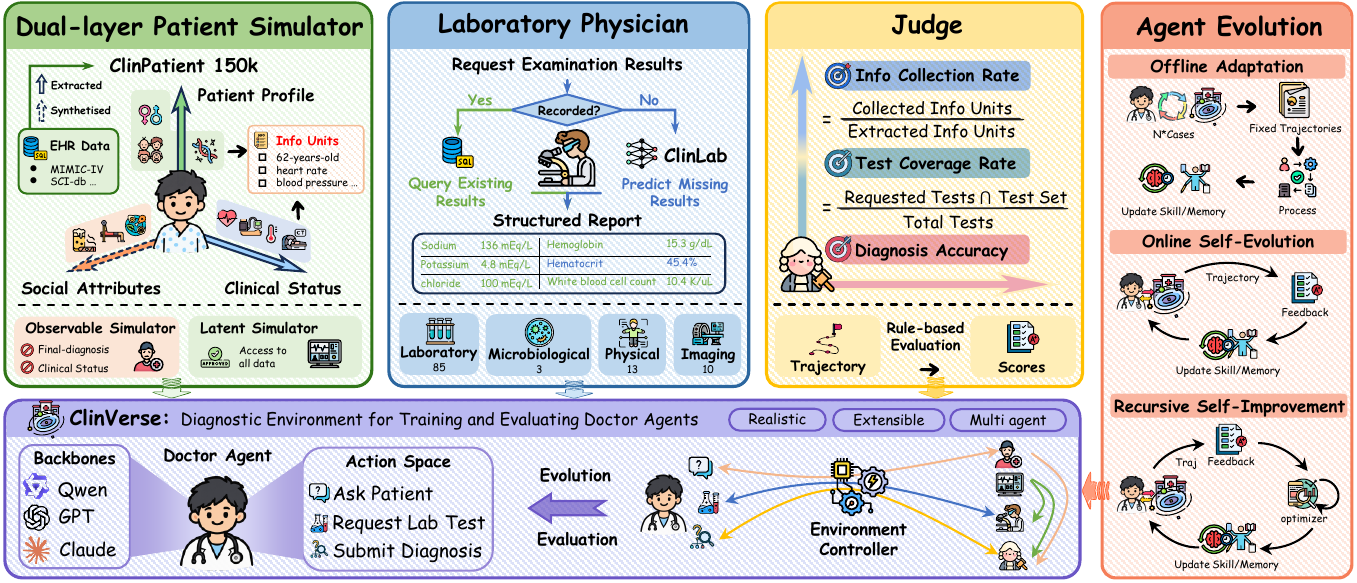}
    \caption{An overview of \textsc{ClinVerse}. Built upon 150k EHR-derived dataset \textsc{ClinPatient}, a dual-layer patient simulator ensures controlled information disclosure, while the laboratory physician retrieves recorded examination results or predicts missing ones using \textsc{ClinLab}. \textsc{ClinVerse} integrates six specialized roles to support multi-turn open-ended clinical interaction. The resulting interaction trajectories are evaluated using three-dimensional rule-based metrics and further utilized for three levels of agent self-evolution strategies.}
    \label{fig:main}
\end{figure*}

\section{Setup of \textsc{ClinVerse}}

As illustrated in Figure~\ref{fig:main}, \textsc{ClinVerse} is a realistic clinical diagnostic environment grounded in real-world EHR data. It comprises six specialized roles that interact freely throughout the consultation, enabling the doctor agent to progressively gather evidence and produce a final diagnosis. The resulting interaction trajectories are evaluated using three categories of rule-based criteria and can subsequently be utilized for both model assessment and training.

\subsection{Task Definition}

For each patient, we construct an EHR-derived clinical case as
\begin{equation}
    \mathcal{C} = \left(\mathcal{U}, \mathcal{E}, y\right),
\end{equation}
where $\mathcal{U}$ and $\mathcal{E}$ denote the patient's clinical
information units and examination items, respectively, and
$y \in \mathcal{Y}_{\mathrm{ICD}}$ is the ground-truth diagnosis represented
by an ICD-10 code~\footnote{https://en.wikipedia.org/wiki/ICD-10}.

The clinical information unit set is defined as
\begin{equation}
    \mathcal{U} = \{u_1, u_2, \ldots, u_M\},
\end{equation}
where each $u_m$ represents patient-specific clinical information, such as a
symptom, medical history, medication, or physical sign. At the
beginning of a consultation, the doctor is provided with an initial subset
$\mathcal{U}_0 \subseteq \mathcal{U}$, which contains the patient's chief
complaint and other initially disclosed information, while the remaining information units
are hidden and can be obtained through interaction with patients.

The examination item set is defined as
\begin{equation}
    \mathcal{E} = \{e_1, e_2, \ldots, e_N\},
\end{equation}
where each $e_n$ denotes an examination item, such as a laboratory test, vital sign measurement, imaging examination,
or other auxiliary examination. The doctor can retrieve an item $e_n$ by
requesting the corresponding examination.

At turn $t$, the doctor receives the currently collected evidence
\begin{equation}
    x_t = \left(\mathcal{U}_t, \mathcal{E}_t\right),
\end{equation}
where $\mathcal{U}_0 \subseteq \mathcal{U}_t \subseteq \mathcal{U}$ and
$\mathcal{E}_t \subseteq \mathcal{E}$ are the revealed patient information
units and retrieved examination items, respectively. The doctor can query the
patient for additional information, request examinations, or terminate the
interaction by issuing a diagnosis.

The task output is a predicted ICD-10 code:
\begin{equation}
    \hat{y} \in \mathcal{Y}_{\mathrm{ICD}}.
\end{equation}
where the ICD-10 code inherently specifies both the disease category and its
fine-grained subtype.

Overall, the task requires the doctor agent to interact iteratively with the patient to collect information units and request relevant examinations to learn about the patient’s physiological state. Once the agent determines that sufficient information has been collected, it provides a final diagnosis of the patient’s underlying disease.

\subsection{Six Roles in \textsc{ClinVerse}}
\label{sec:six_roles}

To simulate realistic clinical interactions while maintaining controlled information access, \textsc{ClinVerse} comprises six specialized roles: an observable patient simulator, a latent patient simulator, a laboratory physician, a doctor agent, an environment controller, and a diagnostic judge. As illustrated in Figure~\ref{fig:main}, the doctor agent interacts with the patient simulators and laboratory physician through the environment controller, progressively acquiring clinical evidence before submitting a final diagnosis. The diagnostic judge evaluates the resulting interaction trajectory, providing feedback for agent evaluation and subsequent self-evolution.

\textbf{Observable Patient Simulator.}
Existing patient simulators~\citep{PatientSim,SDBench} often grant a single model access to the complete patient record, increasing the risk of information leakage under leading or adversarial queries. To address this issue, we adopt a dual-layer patient simulation architecture that separates patient-observable information from latent clinical states. The observable patient simulator is restricted to patient-reportable information, including demographics, chief complaints, subjective symptoms, medical history, and personality traits, while excluding ground-truth diagnoses, clinician-only records, and unrevealed examination results. A self-check mechanism further verifies each generated response against the patient's accessible information and underlying case evidence, revising or rejecting privileged, unsupported, or premature disclosures. Validly revealed information units are recorded for subsequent trajectory-level evaluation.

\textbf{Latent Patient Simulator.}
Complementing the observable patient simulator, the latent patient simulator maintains the complete EHR-derived case state, including the ground-truth diagnosis, longitudinal medical history, physiological measurements, and examination results. It does not interact directly with the doctor agent but provides case-specific clinical information to authorized roles, particularly the laboratory physician. By separating patient-reportable evidence from privileged clinical records, the dual-layer architecture preserves the partial observability of clinical diagnosis and mitigates information leakage. The effectiveness of this design, together with the self-check mechanism, is evaluated in Section~\ref{sec:patient_sim}.

\textbf{Laboratory Physician.}
The laboratory physician handles physical examination requests, laboratory tests, and imaging examinations issued by the doctor agent. As illustrated in Figure~\ref{fig:main}, each request follows a two-stage process: if the corresponding result is recorded in the original EHR, the laboratory physician retrieves it from the latent patient simulator; otherwise, it invokes \textbf{\textsc{ClinLab}}, a trained physiological-state predictor, to infer a clinically plausible result from the case context and available observations. This design avoids directly returning missingness responses for unrecorded examinations, as in prior environments~\citep{AI_Hospital,MIRA}, thereby supporting flexible examination requests while reducing the risk of exploiting systematic missingness as a diagnostic shortcut. Both recorded and predicted results are standardized into structured reports, including measurements, units, reference ranges, and abnormality indicators when applicable. The training and evaluation of \textsc{ClinLab} are detailed in Appendix~\ref{app:ClinLab} and Section~\ref{sec:eval_clinlab}, respectively.

\textbf{Doctor Agent.}
The doctor agent is the primary role evaluated and trained within \textsc{ClinVerse}. Starting from a limited initial patient presentation, it progressively acquires clinical evidence and formulates diagnostic hypotheses through multi-turn interactions. Its action space comprises three types: \textbf{patient inquiry, examination requests, and final diagnosis submission}. Examination requests cover physical examinations, laboratory tests, microbiological tests, and imaging studies. At each turn, the doctor selects an action based on the accumulated interaction history, reflecting the sequential process of clinical evidence acquisition and diagnostic reasoning. The consultation terminates when the doctor submits a final diagnosis or reaches the interaction limit.

\textbf{Environment Controller.}
The environment controller serves as the central coordinator of clinical interactions, connecting the doctor agent with the supporting roles while enforcing information access boundaries. It validates the doctor's actions, routes patient inquiries to the observable patient simulator and examination requests to the laboratory physician, and returns the corresponding observations to the doctor. Throughout the consultation, it maintains the interaction history and records information disclosure and examination events as hidden metadata for subsequent evaluation. This centralized coordination ensures that the doctor obtains clinical evidence only through authorized interactions, preserving the partial observability and consistency of the diagnostic process.

\textbf{Diagnostic Judge.}
The diagnostic judge evaluates the complete interaction trajectory without participating in the consultation. Using the ground-truth case information and interaction metadata maintained by the environment controller, it applies deterministic scoring rules across three dimensions: patient information collection, examination coverage, and diagnostic accuracy at different levels of disease granularity. These scores provide reproducible feedback on both evidence acquisition and diagnostic outcomes, supporting systematic agent evaluation and subsequent self-evolution. The scoring criteria are detailed in Section~\ref{sec:rule_based_scoring}.

\subsection{Interaction Flow}
\label{sec:interaction_flow}

As illustrated in Figure~\ref{fig:main}, each diagnostic episode begins with the environment controller initializing an EHR-derived case from \textsc{ClinPatient} and providing the doctor agent with a limited initial presentation. At each turn, the doctor selects a clinical action based on the accumulated interaction history, which is validated and routed by the environment controller. Patient inquiries are directed to the observable patient simulator, which generates self-checked responses based on patient-accessible information. Physical examination and test requests are handled by the laboratory physician, which retrieves recorded findings from the latent patient simulator or invokes \textsc{ClinLab} to predict unavailable results. The returned observations are incorporated into the interaction history, allowing the doctor to progressively update its diagnostic reasoning and determine subsequent actions.

\subsection{Rule-based Scoring}
\label{sec:rule_based_scoring}

Given a complete interaction trajectory of $T$ turns, we evaluate agent behavior along three dimensions: patient information collection, examination coverage, and diagnostic agreement with the hidden case label. We design deterministic scoring rules to enable reproducible and scalable evaluation while avoiding the inconsistency of LLM-based judges and the cost of expert annotation. These metrics assess how effectively agents acquire and utilize case-specific evidence, rather than judging the clinical optimality of every possible decision path.

\textbf{Information Collection Rate.}
Let $\mathcal{U}$, whose extraction details are provided in Section~\ref{sec:info_unit}, denote the normalized set of patient-reportable information units extracted during case construction, and let $\mathcal{U}_0 \subseteq \mathcal{U}$ denote the subset already disclosed in the initial presentation. The remaining units,
\begin{equation}
    \mathcal{U}^{\mathrm{ask}}
    =
    \mathcal{U} \setminus \mathcal{U}_0,
\end{equation}
represent the information that can be elicited through subsequent interaction with the observable patient simulator.

Each unit is assigned an importance weight
$w(u) \in \{3,2,1\}$,
corresponding to key, supporting, and background evidence, respectively. Let
$\mathcal{U}^{\mathrm{rev}}_T \subseteq \mathcal{U}^{\mathrm{ask}}$
denote the set of distinct units revealed to the doctor over the current trajectory. We define the information collection rate as
\begin{equation}
    R_{\mathrm{info}}
    =
    \frac{
        \sum_{u \in \mathcal{U}^{\mathrm{ask}}}
        w(u)\,
        \mathbf{1}\!\left[u \in \mathcal{U}^{\mathrm{rev}}_T\right]
    }{
        \sum_{u \in \mathcal{U}^{\mathrm{ask}}} w(u)
    }.
\end{equation}

This metric measures the proportion of undisclosed patient-reportable evidence successfully elicited by the doctor, while assigning greater credit to diagnostically salient information. The weighting scheme emphasizes key symptoms, relevant medical history, medications, and allergies, and prevents the score from being dominated by numerous low-value details. For cases in which $\mathcal{U}^{\mathrm{ask}}$ is empty, we set $R_{\mathrm{info}}=1$ by convention.

\textbf{Test Coverage Rate.}
Let $\mathcal{E}^{\mathrm{test}} \subseteq \mathcal{E}$ denote the set of initially hidden and retrievable laboratory, microbiology, and radiology findings associated with the case. Let $\mathcal{E}^{\mathrm{ret}}_T$ denote the distinct examination items returned by the laboratory physician in response to the doctor's requests. We measure overall and abnormal test coverage as
\begin{equation}
    R_{\mathrm{test}}
    =
    \frac{
        \left|
        \mathcal{E}^{\mathrm{ret}}_T
        \cap
        \mathcal{E}^{\mathrm{test}}
        \right|
    }{
        \left|
        \mathcal{E}^{\mathrm{test}}
        \right|
    },
    \qquad
    R_{\mathrm{abnormal}}
    =
    \frac{
        \left|
        \mathcal{E}^{\mathrm{ret}}_T
        \cap
        \mathcal{E}^{\mathrm{abnormal}}
        \right|
    }{
        \left|
        \mathcal{E}^{\mathrm{abnormal}}
        \right|
    }.
\end{equation}

The metrics are computed over distinct case-supported findings rather than the raw number of orders. Therefore, repeated requests do not increase the score, and unsupported requests do not artificially inflate coverage. Here, $\mathcal{E}^{\mathrm{abnormal}}$ denotes the initially hidden laboratory and microbiology findings marked as abnormal. We set the corresponding score to $1$ when its denominator is empty. Thus, $R_{\mathrm{abnormal}}$ measures whether the doctor surfaces clinically salient abnormal test results.

\textbf{Diagnosis Accuracy.}
To provide partial credit for clinically related predictions while maintaining a standardized and language-independent evaluation, we adopt the ICD-10 coding system and compare the predicted code $\hat{y}$ with the ground-truth code $y$ at two levels of granularity. Let $b(\cdot)$ denote the corresponding ICD-10 block and $c_3(\cdot)$ denote the three-character category. We define
\begin{equation}
    A_{\mathrm{cat}}
    =
    \mathbf{1}\!\left[b(\hat{y}) = b(y)\right],
    \qquad
    A_{\mathrm{sub}}
    =
    \mathbf{1}\!\left[c_3(\hat{y}) = c_3(y)\right].
\end{equation}

Here, $A_{\mathrm{cat}}$ measures whether the predicted diagnosis falls within the correct ICD-10 block, while $A_{\mathrm{sub}}$ evaluates the finer-grained match at the three-character category level. All three types of metrics can be computed efficiently, making them suitable both for scalable model evaluation and as reward signals for model training.

\section{Construction of \textsc{ClinPatient}}
\label{sec:clinpatient_construction}
To support scalable patient simulation in \textsc{ClinVerse}, we model each patient from three complementary dimensions and construct a patient pool \textsc{ClinPatient} through a unified data-processing pipeline that converts raw EHR records into diagnosis-oriented cases. Designed to preserve the fidelity of real-world clinical data while remaining extensible, this pipeline can be applied to additional EHR sources, allowing new patient cases to be continuously incorporated into the environment.

\subsection{From EHRs to a Diagnostic Dataset}

\begin{figure}[t]
    \centering
    \includegraphics[width=\columnwidth]{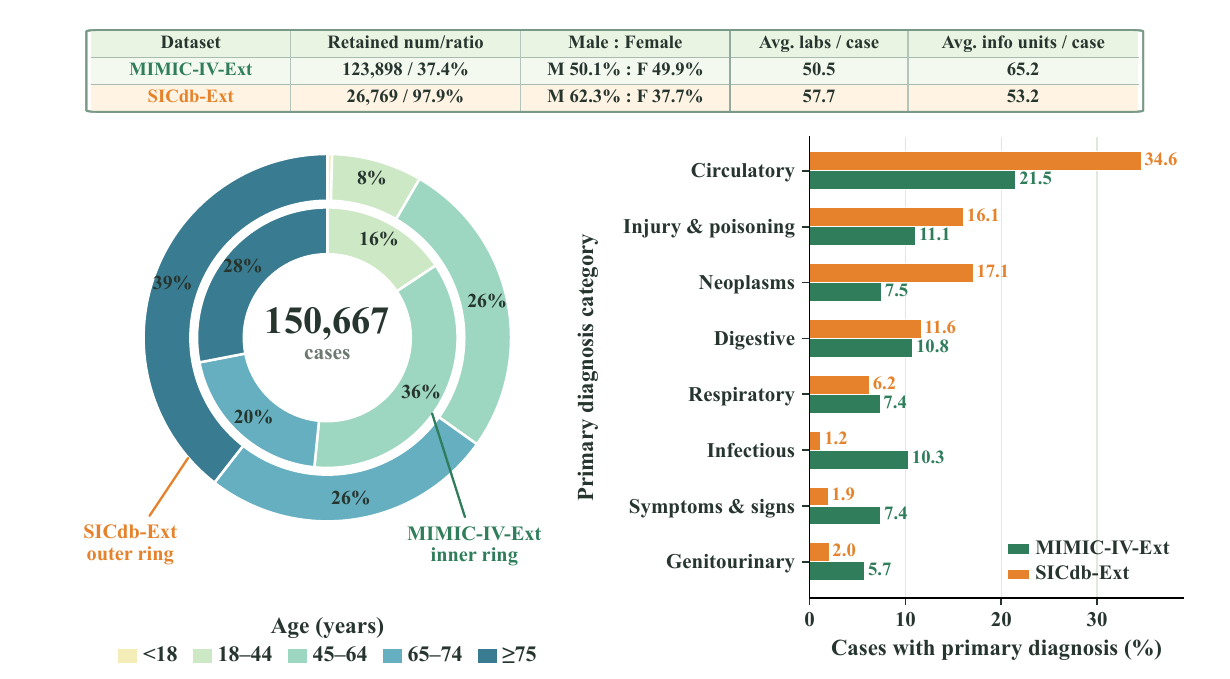}
    \caption{Dataset Statistics of \textsc{ClinPatient}.}
    \label{fig:statistics}
\end{figure}

We construct \textsc{ClinPatient} from structured records in SICdb~\citep{sicdb-1.0.5} and MIMIC-IV~v3.1~\citep{mimiciv-3.1}, together with clinical narratives from MIMIC-IV-Note~\citep{mimic-iv-note-2.2}. These sources are integrated into admission-level datasets, SICdb-Ext and MIMIC-IV-Ext, by aggregating demographics, ICD diagnoses, medications, laboratory and microbiology results, radiology records, and clinical narratives extracted from discharge summaries. We retain only admissions with sufficient evidence, remove diagnosis-revealing text, and preserve only the \emph{Findings} section of radiology reports. The primary diagnosis is normalized to ICD-10 and stored as the hidden ground-truth label $y$, while the processed records provide the acute clinical context and retrievable examination evidence $\mathcal{E}$ without exposing clinician-only information to the observable patient simulator. The statistics of \textsc{ClinPatient} are shown in the Figure~\ref{fig:statistics}.

\begin{table*}[t]
\centering
\caption{Performance comparison of different doctor models and training methods.
Results are grouped by four levels and reported as mean
$\pm$ standard deviation (\%). Methods prefixed with ``+'' are applied to Qwen3-32B;
bold denotes the best result among these adaptation methods. GPT-5.6-Sol with full
information is included as a performance reference.}
\label{tab:main_results}
\resizebox{\textwidth}{!}{
\begin{tabular}{lccccc}
\toprule
\textbf{Doctor Agent}
& \textbf{\makecell{Info Collection\\Rate (\%)}}
& \textbf{\makecell{Test Order\\Coverage (\%)}}
& \textbf{\makecell{Abnormal Test\\Coverage (\%)}}
& \textbf{\makecell{Diagnosis Category\\Accuracy (\%)}}
& \textbf{\makecell{Diagnosis Subtype\\Accuracy (\%)}} \\
\midrule

\rowcolor{gray!15}
\multicolumn{6}{c}{\textbf{L0: Static Agents}} \\
\addlinespace[2pt]

Lingshu-32B
& $28.82 \pm 2.69$
& $0.58 \pm 0.18$
& $1.32 \pm 0.29$
& $28.20 \pm 2.17$
& $11.60 \pm 1.95$ \\

Baichuan-M2-32B
& $26.90 \pm 0.72$
& $56.30 \pm 1.92$
& $65.60 \pm 2.13$
& $29.41 \pm 3.40$
& $19.22 \pm 4.02$ \\

HealthGPT-Pro-4B
& $12.79 \pm 1.87$
& $22.04 \pm 1.18$
& $24.24 \pm 1.60$
& $25.67 \pm 3.06$
& $9.00 \pm 1.73$ \\

HealthGPT-Pro-8B
& $10.62 \pm 1.41$
& $33.90 \pm 2.73$
& $36.56 \pm 3.43$
& $24.00 \pm 8.34$
& $11.60 \pm 5.46$ \\

GPT-4o-mini
& $27.49 \pm 0.41$
& $47.45 \pm 2.42$
& $60.98 \pm 2.64$
& $32.20 \pm 1.10$
& $19.20 \pm 2.39$ \\

GPT-5.6-Luna
& $35.55 \pm 1.29$
& $56.63 \pm 2.06$
& $68.01 \pm 2.19$
& $40.37 \pm 1.92$
& $32.89 \pm 1.68$ \\

Claude-Sonnet-4.6
& $33.45 \pm 1.06$
& $55.93 \pm 1.90$
& $67.42 \pm 2.13$
& $38.07 \pm 1.87$
& $24.19 \pm 2.29$ \\

Qwen3-8B
& $14.33 \pm 1.07$
& $40.70 \pm 1.83$
& $48.44 \pm 2.43$
& $23.20 \pm 2.39$
& $7.80 \pm 2.77$ \\

Qwen3-32B
& $13.96 \pm 1.34$
& $40.84 \pm 1.85$
& $48.06 \pm 1.80$
& $26.04 \pm 2.69$
& $12.43 \pm 1.12$ \\

\midrule
\rowcolor{gray!15}
\multicolumn{6}{c}{\textbf{L1: Offline Adaptation}} \\
\addlinespace[2pt]

\quad + Trace2Skill
& $14.48 \pm 0.93$
& $38.26 \pm 3.64$
& $49.09 \pm 4.48$
& $29.20 \pm 2.49$
& $16.60 \pm 4.45$ \\

\quad + SkillRL
& $15.46 \pm 0.71$
& $37.08 \pm 1.41$
& $47.10 \pm 2.49$
& $27.60 \pm 2.51$
& $16.40 \pm 3.65$ \\

\quad + MemRL-offline
& $14.36 \pm 1.00$
& $43.02 \pm 1.79$
& $54.16 \pm 1.26$
& $31.40 \pm 4.16$
& $16.80 \pm 3.03$ \\

\quad + MedAgent-Zero
& $12.48 \pm 0.65$
& $25.00 \pm 2.06$
& $32.92 \pm 2.72$
& $29.25 \pm 3.77$
& $13.00 \pm 2.58$ \\

\midrule
\rowcolor{gray!15}
\multicolumn{6}{c}{\textbf{L2: Online Self-Evolution}} \\
\addlinespace[2pt]

\quad + MemRL-online
& $\mathbf{15.50 \pm 8.96}$
& $40.34 \pm 1.93$
& $49.65 \pm 2.71$
& $30.87 \pm 4.62$
& $19.33 \pm 3.95$ \\

\quad + Dynamic Cheatsheet
& $11.48 \pm 0.76$
& $\mathbf{49.33 \pm 0.66}$
& $\mathbf{57.45 \pm 1.55}$
& $\mathbf{33.40 \pm 5.86}$
& $\mathbf{20.60 \pm 4.83}$ \\

\quad + ACE
& $13.48 \pm 1.38$
& $39.83 \pm 2.77$
& $51.63 \pm 3.17$
& $30.40 \pm 2.30$
& $20.00 \pm 3.16$ \\

\midrule
\rowcolor{gray!15}
\multicolumn{6}{c}{\textbf{L3: Recursive Self-Improvement}} \\
\addlinespace[2pt]

\quad + Self-Taught Optimizer
& $12.89 \pm 1.16$
& $40.32 \pm 0.97$
& $50.85 \pm 4.07$
& $29.40 \pm 6.11$
& $17.60 \pm 4.98$ \\

\quad + Meta Context Engineering
& $12.84 \pm 1.74$
& $42.11 \pm 2.19$
& $51.43 \pm 3.12$
& $28.40 \pm 5.32$
& $16.40 \pm 4.22$ \\

\quad + MedRSI
& $13.26 \pm 0.36$
& $39.88 \pm 3.14$
& $49.38 \pm 4.98$
& $31.40 \pm 4.51$
& $17.80 \pm 4.21$ \\

\midrule
\addlinespace[2pt]

\rowcolor{gray!15} \textbf{Reference} & \multicolumn{3}{c}{GPT-5.6-Sol + Full Information} & $\mathbf{54.60}$ & $\mathbf{45.80}$ \\

\bottomrule
\end{tabular}
}
\end{table*}

\subsection{Data Augmentation}

To ensure the realism of the environment, we model patients along three complementary dimensions, including patient profile, clinical status, and social attributes. Because real-world admission records often lack social attributes of patients, we augment each case with a conditionally generated Synthea\footnote{https://github.com/synthetichealth/synthea} profile containing patient-reportable information, including prior conditions, long-term medications, allergies, immunization history, and social history. Generation is conditioned on age, sex, and secondary ICD diagnoses, while excluding the primary diagnosis to prevent label leakage. Inconsistent profiles and events occurring after admission or revealing the diagnosis are removed. Each case is also assigned a Big Five personality profile along the OCEAN dimensions to control conversational behaviors such as information disclosure, response conciseness, and anxiety expression~\citep{mccrae1992introduction}. Personality affects only dialogue generation, while all clinical observations and diagnostic targets remain grounded in the original EHR, enabling richer patient simulation without introducing diagnostic shortcuts.

\subsection{Information Unit Extraction}\label{sec:info_unit}

To support controlled information disclosure and trajectory-level evaluation,
we transform long content into a structured set of atomic
information units, following prior work \citep{MedConsultBench}. Specifically, we extract facts from demographics, the history of present illness, and other patient-related contents, and normalize it into a set $\mathcal{U}$. Each unit is assigned a semantic category, an importance label, and a stable identifier. The subset revealed in the initial presentation is denoted by $\mathcal{U}_0 \subseteq \mathcal{U}$, while the remaining units must be acquired through interaction with the observable patient simulator. Examination-derived evidence, including physical examination, laboratory, microbiology, and radiology findings, is stored separately in $\mathcal{E}$ and accessed through the laboratory physician. This separation preserves role-specific information boundaries and provides structured supervision for the scoring functions in Section~\ref{sec:rule_based_scoring}.

\section{Experiments}
We envision \textsc{ClinVerse} as an interactive clinical consultation environment that supports both the evaluation and training of medical agents. In this section, we empirically assess its effectiveness for training and evaluation through a series of experiments.

\begin{figure*}[t]
    \centering
    \includegraphics[width=\textwidth]{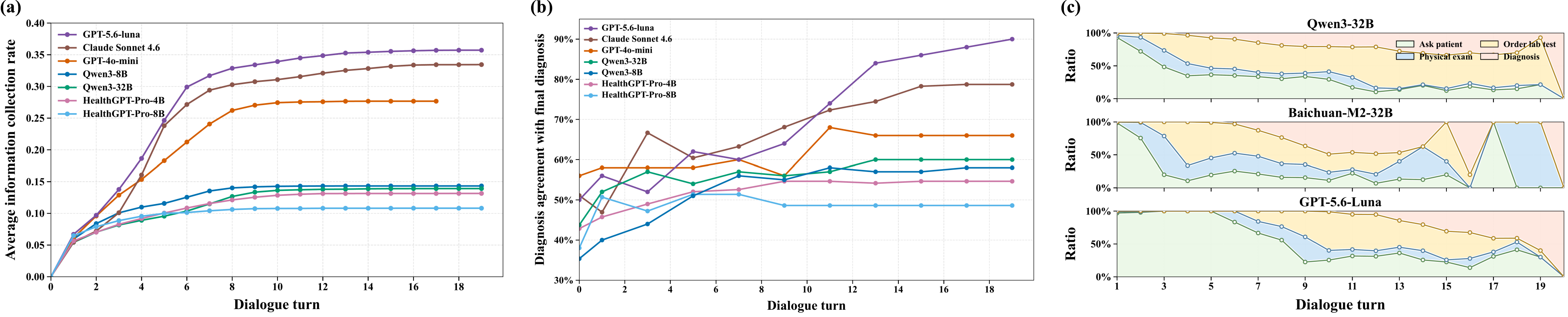}
    \caption{(a) Average information collection rate across dialogue turns. Information is acquired rapidly early in the consultation and gradually plateaus; (b) Average agreement between turn-level and final diagnoses, which increases as more evidence is collected; (c) Distribution of the four action types across turns, showing a general progression from patient inquiry and examination to testing and final diagnosis.
}
    \label{fig:interaction_dynamics}
\end{figure*}

\subsection{Experimental Setup}

\textbf{Metrics.} We evaluate doctor agents using five rule-based metrics introduced in Section~\ref{sec:rule_based_scoring}. All metrics are computed on 100 cases randomly sampled from a held-out test set with 5 independent runs.

\textbf{Models.} We evaluate doctor agents from diverse model families and parameter scales, including general-purpose and medical-domain models such as Lingshu-32B~\citep{Lingshu}, Baichuan-M2-32B~\citep{baichuan-m2}, and the HealthGPT-Pro series~\citep{HealthGPT}. We also include a full-information setting in which complete patient information and examination results are provided directly for agents, whose results are presented in Appendix~\ref{app:full_info}. This setting removes information-access constraints and serves as a reference for interpreting agents' evidence-acquisition and diagnostic capabilities.

\textbf{Training Methods.} We categorize self-evolution methods into three levels. L1 is offline adaptation, including MemRL-offline~\citep{MemRL}, SkillRL~\citep{SkillRL}, Trace2Skill\citep{Trace2Skill},   and MedAgent-Zero~\citep{Agent_Hospital}. L2 represents online self-evolution, including MemRL-online~\citep{MemRL}, Dynamic Cheatsheet~\citep{Dynamic_Cheatsheet}, and ACE~\citep{ACE}. L3 focuses on recursive self-improvement, including Self-Taught Optimizer~\citep{STOP}, Meta Context Engineering~\citep{MCE}, and MedRSI~\citep{medrsi}. Detailed descriptions are provided in Appendix~\ref{app:training_methods}.

\subsection{\textsc{ClinVerse} as an Evaluation Platform}

As presented in Table~\ref{tab:main_results}, we evaluate agents' performance across the full consultation process, covering patient information elicitation, examination ordering, and evidence integration for final diagnosis. GPT-5.6-Luna achieves the strongest overall performance among the evaluated agents, yet a clear gap remains relative to the full-information reference, indicating that effective evidence acquisition is still a major bottleneck in interactive diagnosis. The five metrics further reveal distinct behavioral profiles that cannot be captured by final diagnostic accuracy alone. For example, Baichuan-M2-32B performs well in identifying relevant and abnormal examinations but does not consistently convert this advantage into more accurate diagnoses, suggesting limitations in evidence integration or clinical reasoning. In contrast, Lingshu-32B achieves comparatively strong diagnostic performance despite rarely requesting examinations, implying a greater reliance on prior knowledge or initially available information. By separately evaluating information seeking, examination decisions, and diagnostic inference under controlled information boundaries, \textsc{ClinVerse} exposes where each doctor agent succeeds or fails throughout the consultation process.

\subsection{\textsc{ClinVerse} as an Evolution Environment}

Beyond evaluating static agents, we further investigate how three categories of adaptation and evolution strategies improve Qwen3-32B within \textsc{ClinVerse}. As shown in Table~\ref{tab:main_results}, these methods produce distinct capability profiles. MemRL-online mainly improves Information Collection Rate and Diagnosis Subtype Accuracy, although its information-collection performance exhibits substantial variance, whereas MemRL-offline yields more balanced gains in examination coverage and diagnosis. Dynamic Cheatsheet achieves the strongest overall adaptation results, substantially improving Test Order Coverage, Abnormal Test Coverage, and both diagnostic metrics despite collecting less patient-reported information. This pattern suggests that adaptation can improve diagnosis by shifting evidence acquisition toward more informative examinations rather than uniformly increasing the amount of collected evidence.

All three categories improve diagnostic performance, but their efficiency differs. Online Self-Evolution achieves larger diagnostic gains using 1,000 optimization trajectories than Offline Adaptation using 5,000, indicating substantially greater experience efficiency. Recursive Self-Improvement provides more limited gains under the same setting, suggesting that optimizing the update mechanism introduces additional search and credit-assignment demands that are not necessarily compensated for by the available feedback.

\section{Further Analysis}

\subsection{Info Acquisition \& Diagnostic Convergence}

Figure~\ref{fig:interaction_dynamics}(a--b) shows that information acquisition is concentrated in the early consultation turns and then exhibits clear diminishing returns, whereas diagnostic agreement generally continues to increase as the interaction proceeds. Stronger models achieve both higher information coverage and more consistent diagnoses, suggesting that effective evidence acquisition facilitates diagnostic convergence. Notably, agreement can continue improving after information collection has largely plateaued, indicating that later turns may help models confirm and integrate existing evidence rather than acquire new facts. The occasional fluctuations in agreement further reflect evidence-driven revision of intermediate hypotheses during the consultation.

\subsection{Temporal Analysis of Clinical Actions}

To analyze interaction dynamics, we visualize the turn-level action distributions of three doctor agents. As shown in Figure~\ref{fig:interaction_dynamics}(c), all models exhibit a common coarse-grained workflow: early turns focus on patient inquiry, middle turns increasingly involve physical examinations and laboratory tests, and final diagnoses dominate toward the end. However, the transitions vary across models. GPT-5.6-Luna and Qwen3-32B shift relatively smoothly between action types, whereas Baichuan-M2-32B shows more abrupt changes, suggesting less stable long-horizon planning. These results show that \textsc{ClinVerse} captures not only final performance but also the temporal organization of clinical decision-making.

\subsection{Cross-Model Learning}

Building on Table~\ref{tab:memrl_traj_results}, we further examine how trajectory source and quality affect different target models. Trajectories generated by Qwen3-32B improve Qwen3-series models but provide little benefit to stronger GPT-series models and can even degrade performance, especially for GPT-4o-mini. Reducing the number of trajectories also weakens the gains, indicating that quantity still matters when the teacher remains fixed. However, trajectory quality is more influential: a much smaller set generated by GPT-5.6-Luna yields substantially larger improvements on Qwen3 models than a larger set from Qwen3-32B. These results suggest that trajectory-based training depends strongly on the capability gap between teacher and target models: stronger teachers provide effective supervision for weaker targets, whereas trajectories from weaker or comparable models offer limited value and may introduce harmful guidance.

\definecolor{midred}{HTML}{E43636}
\definecolor{midgreen}{RGB}{34,139,34}

\begin{table}[t]
\centering
\caption{Performance variation of models training with MemRL using trajectories from different models.
All results are reported in percentage (\%) and the five metrics are abbreviated forms of these reported in Table~\ref{tab:main_results}. Cross-model adaptation depends strongly on the source–target capability relationship, with stronger-model trajectories often providing greater benefits despite smaller data volumes.}
\label{tab:memrl_traj_results}

\setlength{\tabcolsep}{2.2pt}
\renewcommand{\arraystretch}{1.15}

\resizebox{\columnwidth}{!}{
\begin{tabular}{llcccccc}
\toprule
\textbf{\makecell{Traj\\Model}}
& \textbf{\makecell{Traj\\Num}}
& \textbf{\makecell{Doctor\\Model}}
& \textbf{\makecell{IC}}
& \textbf{\makecell{TOC}}
& \textbf{\makecell{ATC}}
& \textbf{\makecell{DCA}}
& \textbf{\makecell{DSA}} \\
\midrule

\multirow{5}{*}{Qwen3-32B}
& 5000
& GPT-4o-mini
& ${\textcolor{midgreen}{\downarrow0.30}}$
& ${\textcolor{midgreen}{\downarrow6.16}}$
& ${\textcolor{midgreen}{\downarrow11.41}}$
& ${\textcolor{midgreen}{\downarrow8.20}}$
& ${\textcolor{midgreen}{\downarrow4.15}}$ \\

& 5000
& GPT-5.6-Luna
& ${\textcolor{midgreen}{\downarrow1.86}}$
& ${\textcolor{midgreen}{\downarrow1.94}}$
& ${\textcolor{midgreen}{\downarrow0.65}}$
& ${\textcolor{midgreen}{\downarrow1.31}}$
& ${\textcolor{midred}{\uparrow1.49}}$ \\

& 5000
& Qwen3-8B
& ${\textcolor{midred}{\uparrow0.86}}$
& ${\textcolor{midgreen}{\downarrow2.55}}$
& ${\textcolor{midred}{\uparrow2.65}}$
& ${\textcolor{midred}{\uparrow0.97}}$
& ${\textcolor{midred}{\uparrow3.77}}$  \\

& 5000
&  Qwen3-32B
& ${\textcolor{midred}{\uparrow0.40}}$
& ${\textcolor{midred}{\uparrow2.18}}$
& ${\textcolor{midred}{\uparrow6.10}}$
& ${\textcolor{midred}{\uparrow5.36}}$
& ${\textcolor{midred}{\uparrow4.37}}$  \\

& 500
& Qwen3-32B
& ${\textcolor{midgreen}{\downarrow0.55}}$
& ${\textcolor{midred}{\uparrow1.60}}$
& ${\textcolor{midred}{\uparrow3.79}}$
& ${\textcolor{midred}{\uparrow0.52}}$
& ${\textcolor{midred}{\uparrow4.44}}$  \\
\midrule

\multirow{4}{*}{GPT-5.6-Luna}
& 500
& GPT-4o-mini
& ${\textcolor{midred}{\uparrow4.24}}$
& ${\textcolor{midred}{\uparrow0.86}}$
& ${\textcolor{midred}{\uparrow1.32}}$
& ${\textcolor{midred}{\uparrow1.80}}$
& ${\textcolor{midred}{\uparrow11.80}}$ \\

& 500
& GPT-5.6-Luna
& ${\textcolor{midgreen}{\downarrow0.56}}$
& ${\textcolor{midgreen}{\downarrow2.94}}$
& ${\textcolor{midgreen}{\downarrow1.41}}$
& ${\textcolor{midred}{\uparrow4.73}}$
& ${\textcolor{midred}{\uparrow4.36}}$ \\

& 500
& Qwen3-8B
& ${\textcolor{midred}{\uparrow8.19}}$
& ${\textcolor{midred}{\uparrow8.71}}$
& ${\textcolor{midred}{\uparrow14.07}}$
& ${\textcolor{midred}{\uparrow0.80}}$
& ${\textcolor{midred}{\uparrow8.31}}$  \\

& 500
& Qwen3-32B
& ${\textcolor{midred}{\uparrow6.72}}$
& ${\textcolor{midred}{\uparrow0.87}}$
& ${\textcolor{midred}{\uparrow5.27}}$
& ${\textcolor{midred}{\uparrow5.79}}$
& ${\textcolor{midred}{\uparrow6.37}}$  \\

\bottomrule
\end{tabular}
}
\end{table}

\definecolor{midred}{HTML}{E43636}
\definecolor{midgreen}{RGB}{34,139,34}

\begin{table}[h]
\centering
\caption{Generalization performance of Qwen3-32B adapted with different training methods using interaction trajectories from ClinVerse and evaluated on AIHospital.}
\label{tab:generalization}
\resizebox{\columnwidth}{!}{%
\begin{tabular}{lcccc}
\toprule
\textbf{Doctor Agent}
& \textbf{MVME}
& \textbf{Recall}
& \textbf{Precision}
& \textbf{F1} \\
\midrule

\rowcolor{gray!15}
Qwen3-32B
& 45.67
& 18.81
& 30.42
& 23.25 \\

\quad + MemRL-offline
& 47.40$_{\textcolor{midred}{\uparrow1.73}}$
& 20.41$_{\textcolor{midred}{\uparrow1.60}}$
& 32.78$_{\textcolor{midred}{\uparrow2.36}}$
& 25.16$_{\textcolor{midred}{\uparrow1.91}}$ \\

\quad + SkillRL
& 44.20$_{\textcolor{midgreen}{\downarrow1.47}}$
& 19.69$_{\textcolor{midred}{\uparrow0.88}}$
& 31.28$_{\textcolor{midred}{\uparrow0.86}}$
& 24.17$_{\textcolor{midred}{\uparrow0.92}}$ \\

\quad + MedAgent-Zero
& 45.73$_{\textcolor{midred}{\uparrow0.06}}$
& 18.93$_{\textcolor{midred}{\uparrow0.12}}$
& 31.47$_{\textcolor{midred}{\uparrow1.05}}$
& 23.55$_{\textcolor{midred}{\uparrow0.30}}$ \\

\quad + Trace2Skill
& 45.13$_{\textcolor{midgreen}{\downarrow0.54}}$
& 22.94$_{\textcolor{midred}{\uparrow4.13}}$
& 35.89$_{\textcolor{midred}{\uparrow5.47}}$
& 27.99$_{\textcolor{midred}{\uparrow4.74}}$ \\

\quad + Dynamic Cheatsheet
& 39.45$_{\textcolor{midgreen}{\downarrow6.22}}$
& 26.12$_{\textcolor{midred}{\uparrow7.31}}$
& 36.85$_{\textcolor{midred}{\uparrow6.43}}$
& 30.57$_{\textcolor{midred}{\uparrow7.32}}$ \\

\quad + STOP
& 42.32$_{\textcolor{midgreen}{\downarrow3.35}}$
& 26.46$_{\textcolor{midred}{\uparrow7.65}}$
& 39.44$_{\textcolor{midred}{\uparrow9.02}}$
& 31.67$_{\textcolor{midred}{\uparrow8.42}}$ \\

\bottomrule
\end{tabular}
}
\end{table}

\subsection{Cross-Environment Generalization}\label{cross-env}

To evaluate whether ClinVerse-derived experiences transfer beyond the environment, we test the adapted Qwen3-32B agents on AIHospital~\citep{AI_Hospital} without further adaptation. As shown in Table~\ref{tab:generalization}, all evolution methods consistently improve most metrics, demonstrating that acquired skills from \textsc{ClinVerse} provide transferable supervision across environments. Moreover, we find that most offline adaptation methods maintain relatively stable MVME scores, an LLM-as-a-judge metric for consultation quality, while improving diagnostic accuracy. In contrast, Online Self-Evolution and RSI substantially reduce MVME while achieving larger diagnostic gains. This is consistent with results in Table~\ref{tab:main_results}, where diagnostic accuracy improves despite limited gains in information acquisition.

One possible explanation is that offline methods rely more on accumulated experience and therefore tend to collect additional information for case matching, whereas Online Self-Evolution and RSI continuously refine their strategies toward more efficient decision paths. These results suggest that different evolution paradigms improve agents along distinct capability dimensions.

\begin{figure}[t]
\centering
\includegraphics[width=\columnwidth]{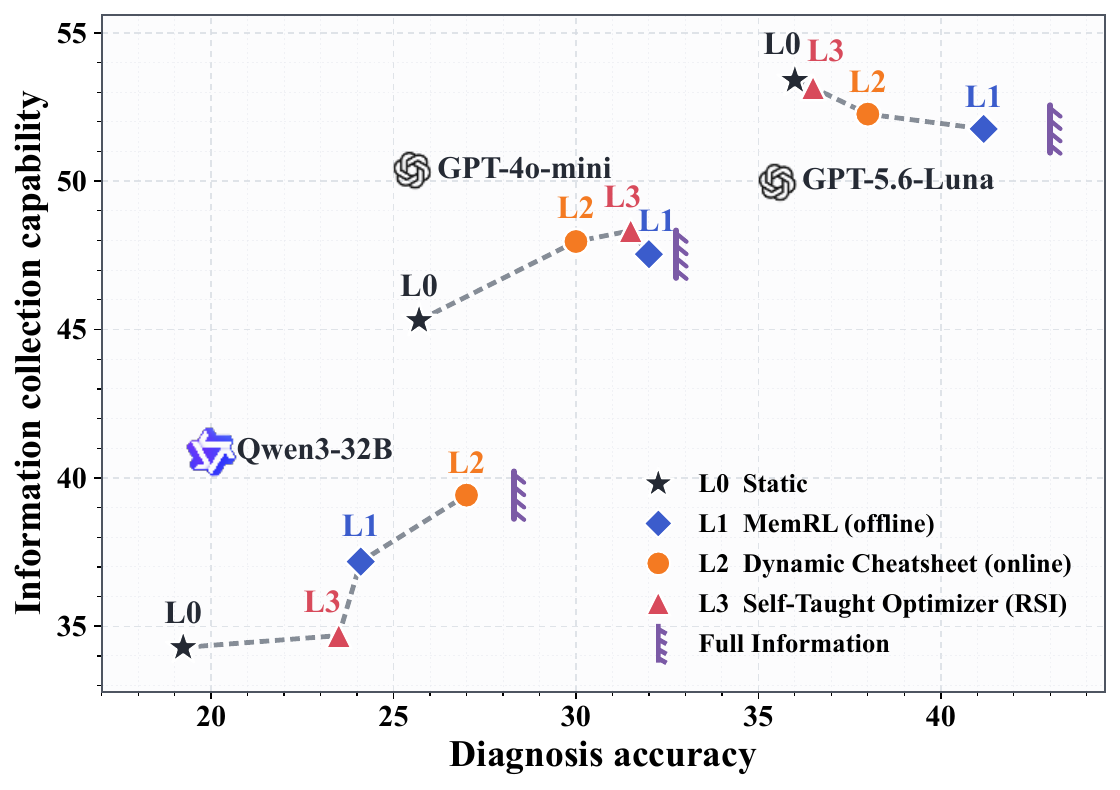}
\caption{Information collection and diagnostic reasoning capabilities across 3 levels of evolution methods and backbone models. Information collection capability is calculated as the average of ICR, TOC, and ATC, while diagnostic accuracy is calculated as the average of DCA and DSA. Greater information acquisition does not necessarily yield better diagnostic accuracy, highlighting the importance of how medical agents select and utilize clinical evidence.}
\label{fig:benefit}
\end{figure}

\subsection{Cross-Methods Improvement}
\label{sec:adaptation}

We further investigate how self-evolution methods affect models of different scales. As shown in Figure~\ref{fig:benefit}, all methods yield positive gains, with performance progressively approaching the full-information reference. Improvements in information collection capability, however, are not always accompanied by gains in diagnostic accuracy. For Qwen3-32B, stronger information collection generally correlates with higher diagnostic accuracy, whereas GPT-5.6-luna can achieve better diagnostic performance with less collected information. This indicates that a substantial portion of patient information may be redundant or weakly relevant to diagnosis, and that self-evolution can help models focus on more informative evidence while reducing unnecessary information acquisition. We also observe a scale-dependent pattern for the L3 RSI method. It provides larger gains for Mid-tier models, while weaker and stronger models benefit less, consistent with the trend reported by~\citet{Harness_Updating}.

\section{Ablation Study}

\subsection{Robustness of Patient Info Boundaries}\label{sec:patient_sim}
Examination results of the dual-layer patient simulator and self-check mechanism described in Section~\ref{sec:six_roles} are presented in Table~\ref{tab:patient_disclosure}, with details in the Appendix~\ref{app:boundary_probe}. The dual-layer design restricts the observable patient simulator from accessing privileged clinical evidence, whereas self-checking filters responses that may disclose information beyond the permitted scope. Their combination achieves the lowest information leakage rate and the highest refusal rate for boundary-violating queries. The results indicate that access control and response verification address distinct sources of leakage and are both necessary for maintaining partial observability.

\begin{table}[t]
\centering
\caption{Comparison of information disclosure and refusal behaviors under different patient simulation settings. }
\label{tab:patient_disclosure}
\resizebox{\columnwidth}{!}{%
\begin{tabular}{lccc}
\toprule
\textbf{Setting}
& \textbf{\makecell{Arbitrary\\Disclosure Rate}}
& \textbf{\makecell{Test-related\\Disclosure Rate}}
& \textbf{Refusal Rate} \\
\midrule

Single\&Neutral
& 47.2\%
& 87.0\%
& 10.0\% \\

Dual-layer\&Neutral
& 36.8\%
& 66.0\%
& 19.7\% \\

Single\&Self-check
& 25.5\%
& 67.0\%
& 62.8\% \\

\rowcolor{gray!15}
Dual-layer\&Self-check
& \textbf{5.7\%}
& \textbf{13.5\%}
& \textbf{84.7\%} \\

\bottomrule
\end{tabular}%
}
\end{table}

\begin{table}[t]
\centering
\caption{Performance comparison across different models as physiological-state predictor. Our trained model ClinLab accurately predicts examination outcomes while achieving high recall for abnormal findings.
}
\label{tab:model_comparison}
\setlength{\tabcolsep}{3.2pt}
\renewcommand{\arraystretch}{1.08}
\resizebox{\columnwidth}{!}{%
\begin{tabular}{lccccc}
\toprule
\textbf{Model}
& \textbf{MAPE}
& \textbf{\makecell{Median\\MAPE}}
& \textbf{MAE}
& \textbf{NormRefErr}
& \textbf{\makecell{Anomaly\\Recall}} \\
\midrule

Qwen3.5-2B
& 57.36
& 27.17
& 77.21
& 777.23
& 30.68 \\

Qwen3.5-27B
& 35.33
& 11.54
& 22.45
& 2.01
& 59.69 \\

DeepSeek-V4-Flash
& 35.47
& 12.50
& 61.63
& 1.92
& 63.95 \\

GPT-4o-mini
& 44.52
& 18.43
& 57.28
& 4.73
& 69.04 \\

\rowcolor{gray!15}
ClinLab-2B
& \textbf{19.46}
& \textbf{3.70}
& \textbf{14.62}
& \textbf{0.36}
& \textbf{76.92} \\

\bottomrule
\end{tabular}%
}
\end{table}

\subsection{Fidelity of Physiological-State Prediction}\label{sec:eval_clinlab}

Table~\ref{tab:model_comparison} presents the evaluation results of the physiological-state predictor ClinLab-2B for examinations whose results are absent from the source EHR. Metrics are detailed in Appendix~\ref{app:ClinLab}. ClinLab-2B consistently outperforms the general-purpose baselines in both numerical accuracy and abnormality recall, suggesting that task-specific training is more effective than simply increasing model scale. By generating plausible examination results while retaining clinically relevant abnormalities, the predictor supports flexible test ordering without allowing missingness in the original EHR to become an unintended diagnostic cue.

\section{Conclusion}

We propose \textsc{ClinVerse}, a scalable real-data-grounded multi-agent environment for medical diagnostic agents. Built on \textsc{ClinPatient}, it supports realistic interactions, flexible examinations, and rule-based evaluation of evidence acquisition and diagnosis. Experiments validate the effectiveness of \textsc{ClinVerse} as an evolution environment and its reliability as an evaluation platform.

\section*{Limitations and Ethical Considerations}

This study uses only de-identified clinical data from SICdb~\citep{sicdb-1.0.5}, MIMIC-IV~v3.1~\citep{mimiciv-3.1}, and MIMIC-IV-Note~\citep{mimic-iv-note-2.2}, accessed through PhysioNet under the required credentialing and Data Use Agreement procedures. We used these data solely for research and made no attempt to identify patients or institutions. Raw EHRs were transformed into structured admission-level cases, and doctor agents accessed only de-identified, role-specific information. Any released artifacts will comply with original licenses and will not include raw records or re-identifiable information.

\section*{Generative AI Usage}
Generative AI tools were used only in an assistive capacity during manuscript preparation and implementation. Specifically, LLM were used for language polishing, grammar and typographical-error checking, and assistance with drafting, debugging, and documenting research code. All AI-assisted text and code were reviewed, revised, and verified by the authors.

\bibliographystyle{ACM-Reference-Format}
\bibliography{references}

\appendix

\section{Related Work}
\subsection{Diagnostic Decision-Making Evaluation}

Existing evaluations of medical agents can be broadly divided into static and interactive paradigms. Static benchmarks typically formulate diagnosis as a question-answering task using cases derived from medical licensing examinations~\citep{MedQA,CMExam}, textbooks~\citep{fouda2026psychiatrybench}, or EHRs~\citep{EHRBench}, with answer accuracy serving as the primary metric. These benchmarks are effective for measuring medical knowledge and diagnostic reasoning under fixed evidence, but they provide limited insight into how agents acquire information or make sequential clinical decisions. More recent benchmarks~\citep{xu2025medagentgym,Agent_Hospital,lim2025matrix} therefore introduce multi-turn consultation settings in which agents interact with simulated patients and other clinical roles, enabling evaluation of both diagnostic outcomes and the decision processes that produce them.

Evaluating such interaction trajectories, however, remains challenging. MIRA~\citep{MIRA} relies on case-specific rules and expert review to assess clinically meaningful behaviors, achieving high evaluation fidelity at substantial annotation cost. Other systems~\citep{AI_Hospital,xu2025medagentgym} improve scalability by using LLM-based judges to score interactions across multiple dimensions, but their results may vary with the evaluator model, prompt design, and scoring criteria. In contrast, \textsc{ClinVerse} adopts a rule-based evaluation framework that derives reproducible scores directly from hidden case states and recorded interaction events, covering information collection, examination ordering, and diagnostic accuracy without requiring per-case expert annotation or model-based judging.

\subsection{Interactive Clinical Environments}

Recent studies~\citep{PhysicianBench,MedAgentSim,SepsisAgent,society,sirry,mu2026ehrworld,li2024mediq} have explored interactive clinical environments for evaluating medical agents beyond static question answering. Agent Hospital~\citep{Agent_Hospital} constructs a flexible hospital sandbox by synthesizing patients and clinical roles from external knowledge sources, but its fully synthetic design may compromise clinical fidelity. In contrast, AIHospital~\citep{AI_Hospital} and SDBench~\citep{SDBench} ground patient simulation in cases collected from online resources or medical literature, providing more realistic clinical scenarios while remaining limited by the scale, completeness, and examination coverage of the source cases.
AgentClinic~\citep{agentclinic} introduces a multimodal clinical environment with four interacting roles, enabling patient consultations, examination requests, and diagnostic decision-making across diverse clinical scenarios. However, its text-parsed interaction protocol and reliance on predefined case findings, with potentially normal responses for unavailable examinations, limit structured clinical interactions and examination flexibility. MIRA~\citep{MIRA} further improves evaluation fidelity through expert-curated cases and fine-grained, case-specific assessment criteria, but the associated annotation burden limits scalability.

Overall, existing environments face a trade-off among interaction flexibility, clinical grounding, and scalable evaluation. \textsc{ClinVerse} differs from these systems by constructing interactive cases from real-world EHRs at scale, providing structured tool-call-based clinical actions, and supporting examination requests beyond the observations explicitly recorded in each source case through ClinLab-based prediction. These designs enable broader and more consistent diagnostic interactions while facilitating subsequent agent self-improvement.

\begin{table}[t]
    \centering
    \caption{Performance of Qwen3-8B before and after SkillRL-GRPO training. Interactive results are reported as mean $\pm$ standard deviation over five runs of 100 cases. The Full Info reference is evaluated on 200 cases and does not involve information collection or test ordering.}
    \label{tab:grpo_results}
    \small
    \setlength{\tabcolsep}{3.5pt}
    \renewcommand{\arraystretch}{1.10}
    \resizebox{\columnwidth}{!}{%
    \begin{tabular}{lccccc}
        \toprule
        \textbf{Model} &
        \makecell{\textbf{Info.}\\\textbf{Collection}} &
        \makecell{\textbf{Test}\\\textbf{Order}} &
        \makecell{\textbf{Abnormal}\\\textbf{Test}} &
        \makecell{\textbf{Diag.}\\\textbf{Category}} &
        \makecell{\textbf{Diag.}\\\textbf{Subtype}} \\
        \midrule
        \rowcolor{gray!15}
        Qwen3-8B
        & $\mathbf{14.3 \pm 1.1}$
        & $40.7 \pm 1.8$
        & $48.4 \pm 2.4$
        & $23.2 \pm 2.4$
        & $7.8 \pm 2.8$ \\
        \quad + SkillRL-GRPO
        & $6.7 \pm 0.2$
        & $\mathbf{45.2 \pm 1.5}$
        & $\mathbf{57.6 \pm 1.7}$
        & $28.6 \pm 2.1$
        & $16.6 \pm 1.1$ \\
        \quad + Full Info
        & -- & -- & --
        & $\mathbf{32.6}$
        & $\mathbf{17.5}$ \\
        \bottomrule
    \end{tabular}%
    }
\end{table}

\section{Training Methods Used in Experiments}\label{app:training_methods}

We adapt doctor agents within \textsc{ClinVerse} using nine methods. Most of the methods keep the backbone LLM frozen and improve agent behavior through external skills or memories derived from interaction trajectories.

\textbf{Trace2Skill}~\citep{Trace2Skill} collects successful and failed clinical trajectories from a frozen doctor agent and analyzes them separately to extract reusable behaviors and causally grounded corrections. These trajectory-level lessons are converted into local skill patches, then hierarchically deduplicated and consolidated into a coherent clinical skill directory. The resulting skill is directly supplied at inference time without model updates or per-case memory retrieval.

\textbf{SkillRL}~\citep{SkillRL} distills successful trajectories into effective decision patterns and failed trajectories into counterfactual lessons describing errors and preferable actions. It organizes this knowledge into a hierarchical \textsc{SkillBank} containing general clinical principles and task-specific guidance for stages such as history collection, investigation, and diagnosis. Our parameter-free adaptation retrieves relevant skills as contextual guidance while omitting the SFT and reinforcement-learning updates of the original framework. We additionally apply the full SkillRL training pipeline to Qwen3-8B, with details and results reported in Appendix~\ref{app:GRPO}.

\textbf{MemRL}~\citep{MemRL} represents each clinical episode as an intent -- experience -- utility triplet in an external memory bank initialized from offline rollouts. At inference time, it first recalls semantically similar experiences and then re-ranks them using learned utility, reducing the influence of relevant-looking but unhelpful memories. Environmental feedback subsequently updates the utilities of retrieved experiences and may add new memories, enabling online adaptation without changing model weights.

\textbf{MedAgent-Zero}~\citep{Agent_Hospital} maintains separate Medical Record Libraries and Experience Bases for the examination, diagnosis, and treatment stages. Successful cases are stored as validated examples, whereas failed cases are reflected upon and summarized into transferable corrective principles. For each new clinical decision, similar records and failure-derived guidance are jointly retrieved to support reasoning, allowing both repositories—and thus the doctor agent—to improve continually without parameter updates.

\textbf{Dynamic Cheatsheet}~\citep{Dynamic_Cheatsheet} gives a black-box LLM a persistent, self-curated memory that accumulates concise strategies, code snippets, and corrections across test-time interactions. After each case, a curator examines the current memory, query, and attempted solution to add useful generalizations and revise or prune unreliable entries; the revised cheatsheet then conditions subsequent cases. This enables online adaptation without ground-truth labels, human feedback, or changes to model parameters, while limiting the context growth caused by appending the full interaction history.

\textbf{ACE}~\citep{ACE} treats the agent context as an evolving playbook and separates adaptation into generator, reflector, and curator roles. The generator acts using the current playbook, the reflector extracts generalizable lessons from the resulting trajectory and natural execution feedback, and the curator integrates compact delta updates before semantically deduplicating overlapping entries. We use its online setting so that later consultations receive the refined playbook; these localized updates preserve useful clinical detail and mitigate the brevity bias and context collapse associated with repeatedly rewriting the entire context.

\textbf{Self-Taught Optimizer}~\citep{STOP} begins with an LLM-powered improver program that proposes multiple revisions to an input program, evaluates them with a task-specific utility function, and returns the best candidate. The seed improver is then applied to its own source code, allowing utility feedback to produce a stronger update procedure whose emergent search strategies can include beam search, genetic algorithms, and simulated annealing. The improved improver is subsequently used to optimize agent solutions, yielding a form of recursive self-improvement in which the scaffolding evolves while the backbone LLM remains frozen.

\textbf{Meta Context Engineering}~\citep{MCE} formulates adaptation as bi-level optimization that co-evolves context-engineering skills and the context artifacts produced by those skills. At the meta level, agentic crossover synthesizes a revised skill from the task specification and the history of prior skills, executions, and evaluations; at the base level, an agent executes that skill to construct context as flexible files and code, including both static resources and query-dependent operations such as retrieval and filtering. Iterative evaluation selects improved skill--context pairs, allowing both what the doctor agent learns and how it organizes and updates that knowledge to evolve without modifying the backbone model.

\textbf{MedRSI}~\citep{medrsi} enables recursive self-improvement by transforming diagnostic failures into new clinical capabilities while keeping the backbone LLM frozen. It prioritizes failures according to their potential clinical consequences and identifies missing capabilities through trajectory analysis and reflection. These capability gaps are addressed through automated code generation, tool composition, or task-specific model training, producing new clinical tools for measurement, prediction, and evidence integration. To ensure reliable adaptation, MedRSI adopts a fast-discovery, slow-registration mechanism, where candidate tools are evaluated across subsequent patient cohorts and incorporated into a persistent tool registry only after demonstrating sustained diagnostic improvements without increasing clinical risk. This iterative process allows the doctor agent to autonomously expand its capabilities without updating the backbone LLM.

\begin{figure}[t]
    \centering
    \includegraphics[width=\columnwidth]{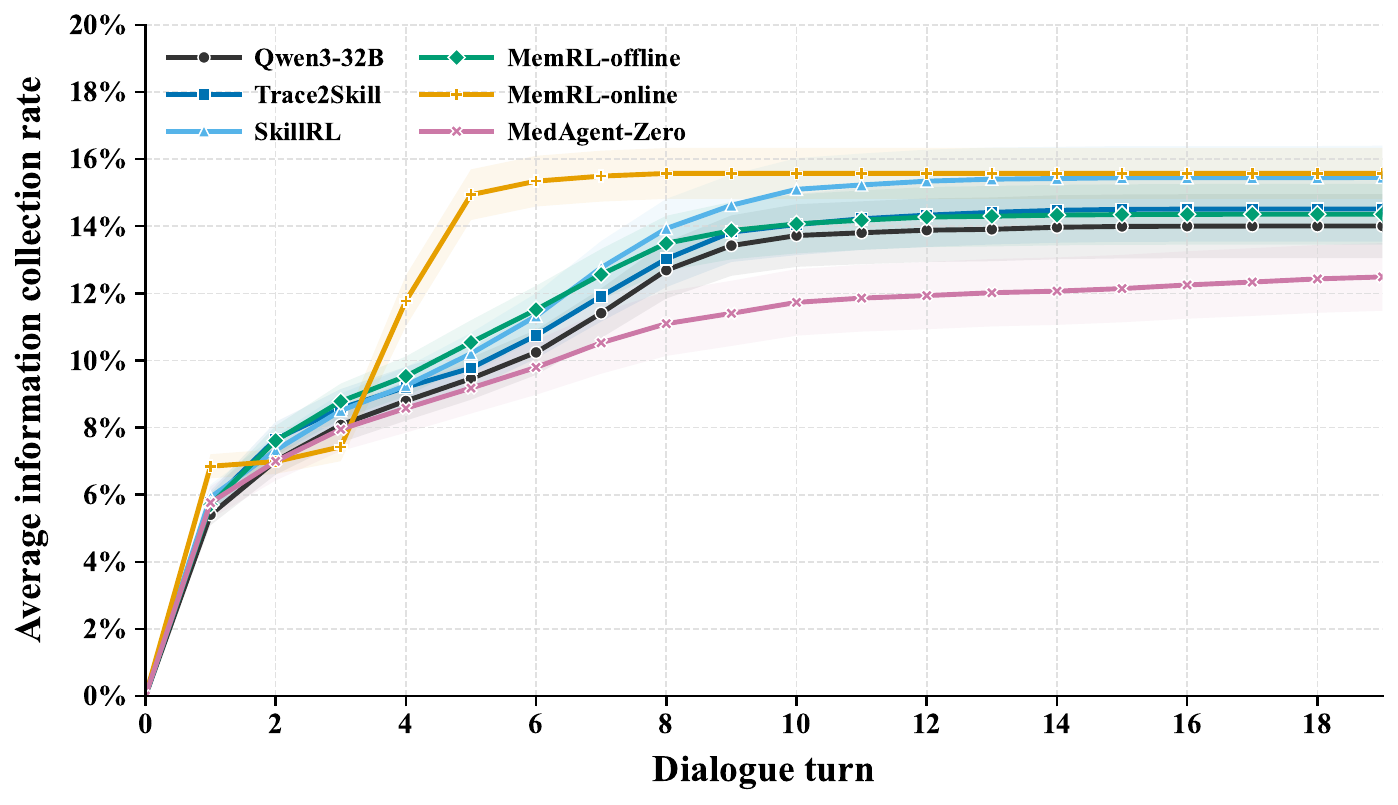}
    \caption{Comparison of Information Collection Rate of Different Training Methods with Qwen3-32B}
    \label{fig:ICR_training}
\end{figure}

\section{Effectiveness of ClinVerse for GRPO Training}\label{app:GRPO}

To further examine whether \textsc{ClinVerse} can support parameter-updating reinforcement learning, we adapt the GRPO stage of \textit{SkillRL} to the \textsc{ClinVerse} interaction and reward interface and optimize an SFT-initialized Qwen3-8B doctor agent for 30 steps. During training, the agent conducts multi-turn clinical encounters and receives the rule-based \textsc{ClinVerse} score as its reward, allowing the policy to learn directly from its information-gathering and diagnostic decisions.

As shown in Table~\ref{tab:grpo_results}, SkillRL-GRPO consistently improves examination decisions and diagnostic performance, indicating that the reward signals provided by \textsc{ClinVerse} can effectively guide policy optimization. Notably, the trained model achieves subtype-level diagnostic performance close to the Full Info reference, despite operating under the original partially observable setting. Meanwhile, the decrease in information collection suggests that the optimized policy relies more on examination-based evidence and diagnostic inference rather than extensive patient inquiry. These results demonstrate that \textsc{ClinVerse} not only provides effective training signals for GRPO but also enables fine-grained analysis of how optimization reshapes clinical decision-making behaviors.

\begin{figure}[t]
    \centering
    \includegraphics[width=\columnwidth]{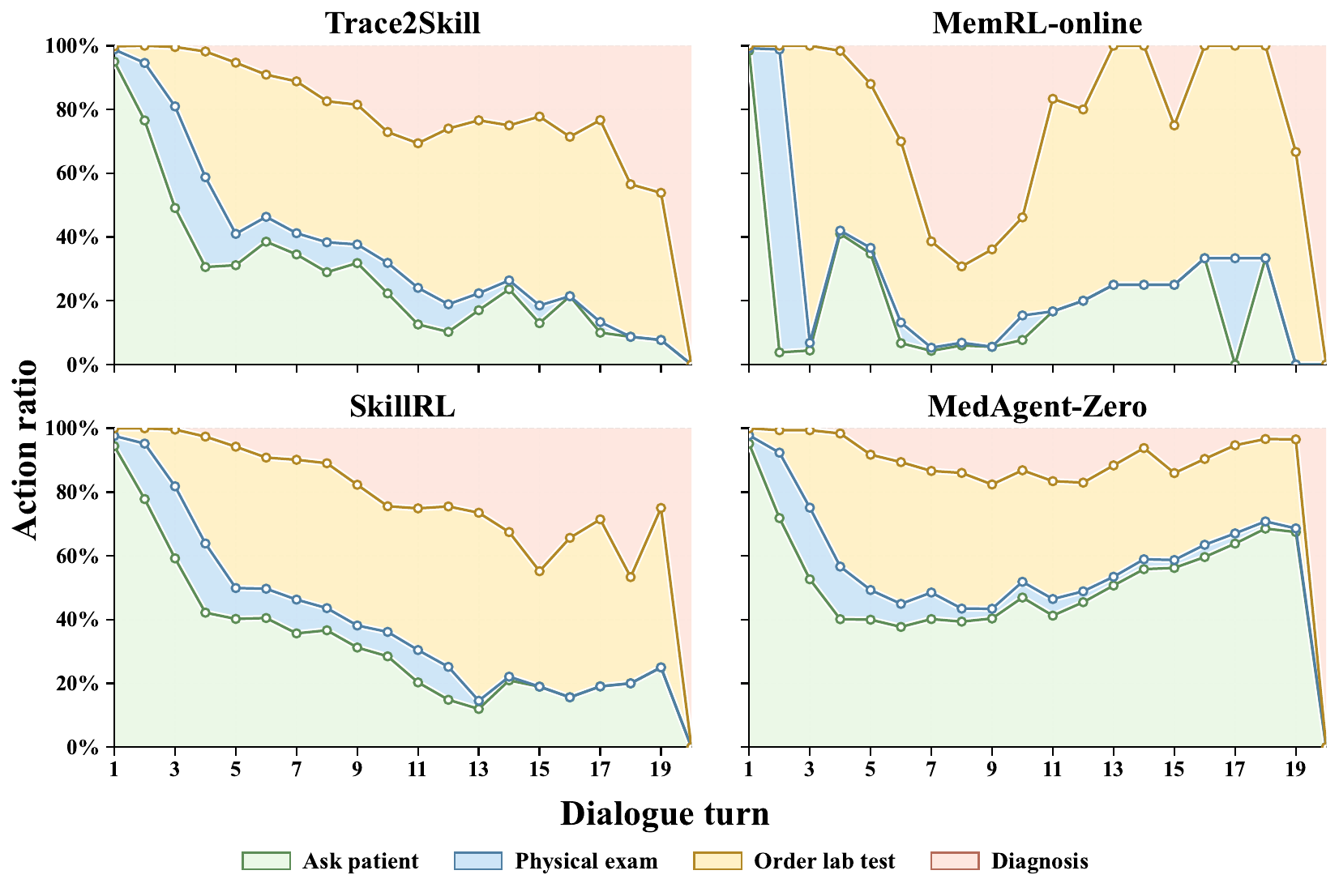}
    \caption{Comparison of Action Taken Ratio of Different Training Methods with Qwen3-32B}
    \label{fig:action_training}
\end{figure}

\section{Impact of Different Training Methods on Model Behavior}

\subsection{Impact on Information Collection}

Figure~\ref{fig:ICR_training} shows that all training methods enable agents to acquire most patient-reportable information during the early stages of consultation, after which the collection rate gradually saturates. However, different methods exhibit distinct acquisition patterns. MemRL-online accelerates early evidence acquisition and reaches a stable level faster, while SkillRL and MemRL-online achieve the strongest overall information collection capability. In contrast, only limited improvements are provided by Trace2Skill and MemRL-offline over the original model, and MedAgent-Zero even reduces information coverage. These results indicate that trajectory-based adaptation influences not only the amount of information collected but also the timing of evidence acquisition. In particular, utility-aware memory retrieval encourages more efficient early-stage questioning, whereas improvements in diagnostic performance do not necessarily imply more comprehensive information gathering.

\subsection{Impact on Action Taken}

The turn-level action distributions in Figure~\ref{fig:action_training} reveal that different training methods induce distinct interaction strategies. Trace2Skill and SkillRL maintain a relatively smooth clinical workflow, gradually transitioning from patient inquiry to examination requests and final diagnosis. MemRL-online produces a more structured and stage-wise interaction pattern, with each action type concentrated within specific phases of the consultation, enabling rapid progression toward diagnosis. In contrast, MedAgent-Zero maintains prolonged patient inquiry throughout the interaction and performs fewer diagnostic actions, suggesting that its retrieved experiences encourage extended but less targeted information gathering. These results show that trajectory-based adaptation not only improves final performance but also reshapes the temporal organization of clinical decision-making, with different methods achieving distinct trade-offs between exploration and efficiency.

\section{Performances of Different Models Under Full Information Setting}\label{app:full_info}

\begin{table}[t]
\centering
\caption{Diagnostic performance of different models under the full-information setting.}
\label{tab:full_info_results}
\begin{tabular}{lcc}
\toprule
\textbf{Doctor Model}
& \textbf{\makecell{Diagnosis Category\\Accuracy (\%)}}
& \textbf{\makecell{Diagnosis Subtype\\Accuracy (\%)}} \\
\midrule

GPT-5.6-Luna
& 48.00
& 38.00 \\

GPT-5.6-Sol
& 54.00
& 45.00 \\

Qwen3-8B
& 32.60
& 17.50 \\

Qwen3-32B
& 35.00
& 21.60 \\

HealthGPT-Pro-4B
& 32.50
& 9.50 \\

HealthGPT-Pro-8B
& 32.40
& 17.80 \\

Lingshu-32B
& 35.20
& 22.60 \\

GPT-4o-mini
& 38.00
& 27.50 \\

\bottomrule
\end{tabular}
\end{table}

Table~\ref{tab:full_info_results} reports diagnostic performance when complete patient information and examination results are directly provided, isolating evidence integration and diagnostic reasoning from information acquisition. Compared with the interactive setting, the ranking of models changes substantially, suggesting that strong diagnostic reasoning with complete evidence does not necessarily translate into effective evidence acquisition and interaction planning. GPT-series models generally benefit more from full information, while Qwen3, HealthGPT-Pro, and Lingshu models show relatively smaller gaps between constrained and unconstrained settings. Model scaling also exhibits different effects across model families, indicating that larger parameter sizes do not uniformly improve fine-grained diagnosis. Moreover, all models show a consistent performance drop from category-level to subtype-level diagnosis, demonstrating that fine-grained clinical reasoning remains challenging even when information access is no longer a limitation.

\section{Details of Cross-Methods Improvement}
Table~\ref{tab:cross_model_training_results} provides the complete numerical results underlying Figure~\ref{fig:benefit} and the cross-method analysis in Section~\ref{sec:adaptation}. While the figure highlights relative changes in information collection and diagnostic accuracy, the table reports the absolute values of all five evaluation metrics for every model--method combination.

\begin{table}[t]
\centering
\caption{Performance of three doctor models under static inference and three adaptation methods. IC, TOC, ATC, DCA, and DSA denote Info Collection, Test Order Coverage, Abnormal Test Coverage, Diagnosis Category Accuracy, and Diagnosis Subtype Accuracy, respectively.}
\label{tab:cross_model_training_results}
\setlength{\tabcolsep}{6pt}
\renewcommand{\arraystretch}{1.12}
\resizebox{\columnwidth}{!}{
\begin{tabular}{llccccc}
\toprule
\textbf{Doctor Model}
& \textbf{Method}
& \textbf{IC}
& \textbf{TOC}
& \textbf{ATC}
& \textbf{DCA}
& \textbf{DSA} \\
\midrule

\multirow{4}{*}{Qwen3-32B}
& Static & 13.96 & 40.84 & 48.06 & 26.04 & 12.43 \\
& MemRL & 14.36 & 43.02 & 54.16 & 31.40 & 16.80 \\
& Dynamic Cheatsheet & 11.48 & 49.33 & 57.45 & 33.40 & 20.60 \\
& Self-Taught Optimizer & 12.89 & 40.32 & 50.85 & 29.40 & 17.60 \\
\midrule

\multirow{4}{*}{GPT-4o-mini}
& Static & 27.49 & 47.45 & 60.98 & 32.20 & 19.20 \\
& MemRL & 28.06 & 50.21 & 64.35 & 38.00 & 26.00 \\
& Dynamic Cheatsheet & 29.36 & 50.44 & 64.10 & 34.00 & 26.00 \\
& Self-Taught Optimizer & 27.12 & 49.65 & 68.22 & 37.00 & 26.00 \\
\midrule

\multirow{4}{*}{GPT-5.6-Luna}
& Static & 35.55 & 56.63 & 68.01 & 40.00 & 32.00 \\
& MemRL & 34.99 & 53.69 & 66.60 & 45.10 & 37.25 \\
& Dynamic Cheatsheet & 29.80 & 55.67 & 71.31 & 41.00 & 35.00 \\
& Self-Taught Optimizer & 32.07 & 54.76 & 72.62 & 41.00 & 32.00 \\

\bottomrule
\end{tabular}
}
\end{table}

\section{Transfer from AIHospital to ClinVerse}
\label{app:aihospital_to_clinverse}

In Section~\ref{cross-env}, we demonstrate that the capabilities acquired through training in ClinVerse can be transferred to AIHospital. We further examine the reverse transfer direction by adapting Qwen3-32B with MemRL, SkillRL, and MedAgent-Zero using trajectories collected from 406 AIHospital~\citep{AI_Hospital} cases, then evaluating the agents in \textsc{ClinVerse}. Table~\ref{tab:aihospital_to_clinverse} reports changes relative to the unadapted agents. Most method show no improvement: MemRL and MedAgent-Zero decline across the reported metrics, while SkillRL improves information collection and examination coverage but yields only small diagnostic gains. These results suggest that adaptation in AIHospital does not consistently translate into improved performance in \textsc{ClinVerse}, and further demonstrate the effectiveness and validity of ClinVerse's design.

\definecolor{midred}{HTML}{E43636}
\definecolor{midgreen}{RGB}{34,139,34}

\begin{table}[t]
\centering
\caption{Changes relative to Baseline for Qwen3-32B agents adapted in AIHospital and evaluated in \textsc{ClinVerse}. Metric abbreviations follow Table~\ref{tab:main_results}. Upward and downward arrows indicate increases and decreases, respectively.}
\label{tab:aihospital_to_clinverse}
\small
\setlength{\tabcolsep}{3pt}
\renewcommand{\arraystretch}{1.15}

\begin{tabular}{lccccc}
\toprule
\textbf{Method} & \textbf{IC} & \textbf{TOC} & \textbf{ATC} & \textbf{DCA} & \textbf{DSA} \\
\midrule
MemRL
& $\textcolor{midgreen}{\downarrow2.24}$
& $\textcolor{midgreen}{\downarrow4.80}$
& $\textcolor{midgreen}{\downarrow4.71}$
& $\textcolor{midgreen}{\downarrow1.60}$
& $\textcolor{midgreen}{\downarrow1.00}$ \\
SkillRL
& $\textcolor{midred}{\uparrow3.23}$
& $\textcolor{midred}{\uparrow2.12}$
& $\textcolor{midred}{\uparrow3.49}$
& $\textcolor{midred}{\uparrow0.40}$
& $\textcolor{midred}{\uparrow1.20}$ \\
MedAgent-Zero
& $\textcolor{midgreen}{\downarrow5.71}$
& $\textcolor{midgreen}{\downarrow15.89}$
& $\textcolor{midgreen}{\downarrow18.29}$
& $\textcolor{midgreen}{\downarrow7.00}$
& $\textcolor{midgreen}{\downarrow4.60}$ \\
\bottomrule
\end{tabular}%
\end{table}

\section{Diagnosis with an Oracle ICD-10 Block}
\label{app:full_info_category}

\definecolor{midred}{HTML}{E43636}
\definecolor{midgreen}{RGB}{34,139,34}

\begin{table}[t]
    \centering
    \caption{Diagnosis subtype accuracy under full information with the ground-truth ICD-10 block provided as a hard constraint. The model must predict the correct three-character category within the given block.}
    \label{tab:full_info_category_results}
    \small
    \setlength{\tabcolsep}{8pt}
    \renewcommand{\arraystretch}{1.10}
    \begin{tabular}{lc}
        \toprule
        \textbf{Doctor Model}
        & \textbf{Diagnosis Subtype Accuracy (\%)} \\
        \midrule
        GPT-4o-mini & 51${\textcolor{midred}{\uparrow31.80}}$ \\
        GPT-5.6-Luna & \textbf{73}${\textcolor{midred}{\uparrow40.11}}$ \\
        Qwen3-32B & 45${\textcolor{midred}{\uparrow32.57}}$ \\
        Lingshu-32B & 49${\textcolor{midred}{\uparrow37.40}}$ \\
        Baichuan-M2-32B & 14${\textcolor{midgreen}{\downarrow5.22}}$ \\
        \bottomrule
    \end{tabular}
\end{table}

We further investigate fine-grained diagnostic reasoning under an oracle
block-conditioned setting, where models are provided with complete patient
information, examination results, and the ground-truth ICD-10 block as a hard
constraint. The task is to predict the corresponding three-character
diagnostic category within the given block, and predictions violating the
constraint are considered incorrect. As shown in Table~\ref{tab:full_info_category_results},
oracle block information substantially improves subtype-level diagnosis across
models, indicating that coarse disease localization accounts for a major
portion of diagnostic errors. Nevertheless, considerable performance gaps
remain between models, demonstrating that fine-grained category selection
requires additional clinical reasoning and constraint-following ability beyond
access to the correct diagnostic region. The consistent ranking with the
unconditioned full-information setting further suggests that stronger models
maintain advantages even when the search space is narrowed by oracle
information.

\section{Training Detail of Physiological-state Predictor}\label{app:ClinLab}

\textbf{Training.}
We construct the training set from laboratory measurements in MIMIC-IV-Ext and split samples by hospital admission to prevent admission-level context leakage between training and evaluation. For each eligible measurement, we apply a leave-one-out masking strategy, where the target laboratory result is removed while all remaining clinical information, including demographics, history, physical examination, diagnoses, medications, microbiology results, and hospital course, is preserved. The model receives the masked test name, category, unit, and reference range, and is trained to predict only the missing result value. This formulation simulates real scenarios where requested examinations are absent from the source EHR while preventing direct target copying. To emphasize clinically meaningful cases, all abnormal measurements are retained and normal measurements are subsampled. We fine-tune a Qwen3.5-2B backbone with LoRA under the causal language modeling objective. During training, system, user, and padding tokens are assigned the ignore index $-100$, ensuring that only the generated laboratory result contributes to the loss. For samples exceeding the context limit, inputs are truncated while preserving the target response.

\textbf{Evaluation Metrics.}
Given the ground-truth value $y_i$ and prediction $\hat{y}_i$, we evaluate numerical accuracy using four metrics. \textit{MAPE} computes the mean absolute percentage error, $\frac{1}{N}\sum_i|\hat{y}_i-y_i|/|y_i|$, while \textit{Median MAPE} reports the median sample-level percentage error to reduce sensitivity to outliers. \textit{MAE} measures the absolute deviation between predictions and ground truth, and \textit{NormRefErr} normalizes the absolute error by the width of the corresponding reference interval, $|\hat{y}_i-y_i|/(u_i-l_i)$, enabling comparison across heterogeneous laboratory tests. We further measure \textit{Anomaly Recall}, which evaluates whether predicted values correctly identify ground-truth abnormal results based on test-specific reference ranges. Lower values indicate better performance for MAPE, Median MAPE, MAE, and NormRefErr, while higher values indicate better performance for Anomaly Recall. Following the evaluation protocol, zero-valued targets are excluded from numerical error metrics, unstable tests with per-test MAPE above 200\% are removed, and individual percentage errors are clipped at 200\% when computing aggregate MAPE.

\section{Information-Boundary Stress Test and Interaction Impact}
\label{app:boundary_probe}

\textbf{Stress-test protocol.}
The ablation in Table~\ref{tab:patient_disclosure} evaluates information-boundary robustness under adversarial queries rather than routine consultation. For each of 100 cases, we issue six boundary-seeking queries covering diagnosis requests, examination-result requests, and attempts to obtain complete medical records. We compare different combinations of information architecture (single simulator vs. dual layer) and response policy (neutral vs. self-check) under identical cases and prompts, resulting in 600 responses per setting.

\textbf{Metric definitions.}
\textit{Arbitrary Disclosure Rate} measures the proportion of responses containing any suspected privileged disclosure, including explicit diagnosis leakage or unsupported objective findings. \textit{Test-related Disclosure Rate} focuses on leakage of examination results among test-related queries. \textit{Refusal Rate} measures responses containing explicit refusal or unavailability statements. These metrics are computed independently, and disclosure labels are obtained through deterministic lexical screening rather than manual clinical judgment.

\textbf{Impact on ordinary interaction.}
The high refusal rate observed under the adversarial protocol reflects the intentionally boundary-violating nature of the queries and does not represent routine consultation behavior. To examine practical impact, we further analyze 98 completed Qwen3-32B trajectories from the main evaluation, containing 393 patient-query turns. Only a small fraction of responses match the refusal lexicon, and most correspond to normal uncertainty about symptoms, history, or medications rather than information-boundary violations. Only two cases involve explicit requests for unavailable diagnostic or examination results, resulting in a negligible boundary-triggered refusal frequency. These cases appropriately redirect the doctor to obtain objective evidence through the laboratory physician.

These results indicate that the proposed boundary mechanism can resist explicit information leakage attempts while minimally affecting normal patient interaction.

\section{ClinLab Validity Analysis}
\label{app:clinlab_validity}

To investigate whether ClinLab-generated examination results affect diagnostic reliability, we conduct two additional experiments examining the robustness of diagnostic performance to predicted values and missing examination results.

\textbf{Robustness to ClinLab predictions.}
We randomly replace 10\% and 20\% of available EHR examination values with ClinLab predictions and evaluate the resulting diagnostic performance. As shown in Table~\ref{tab:clinlab_robustness}, both category and subtype accuracy remain stable across different replacement ratios, suggesting that the introduction of ClinLab-generated values does not substantially compromise overall diagnostic performance.

\begin{table}[t]
    \centering
    \caption{Diagnostic performance under different proportions of ClinLab-predicted examination values.}
    \label{tab:clinlab_robustness}
    \small
    \begin{tabular}{lcc}
        \toprule
        Replacement Ratio & DCA & DSA \\
        \midrule
        0\%  & 26.04 $\pm$ 2.69 & 12.43 $\pm$ 1.12 \\
        10\% & 26.41 $\pm$ 2.30 & 13.22 $\pm$ 1.48 \\
        20\% & 28.27 $\pm$ 1.92 & 14.61 $\pm$ 1.52 \\
        \bottomrule
    \end{tabular}

\end{table}

\textbf{Impact of missing examination results.}
We further compare ClinLab-based prediction with explicitly returning \texttt{null} for unavailable examinations. As shown in Table~\ref{tab:clinlab_missing}, removing ClinLab leads to higher diagnostic accuracy, particularly at the category level. One possible explanation is that explicit missingness provides additional cues about the underlying diagnosis, which agents may exploit rather than relying solely on acquired clinical evidence.

\begin{table}[t]
    \centering
    \caption{Diagnostic performance with and without ClinLab.}
    \label{tab:clinlab_missing}
    \small

    \begin{tabular}{lcc}
        \toprule
        Setting & DCA & DSA \\
        \midrule
        w/ ClinLab  & 35.58 & 31.73 \\
        w/o ClinLab & 39.17 & 31.91 \\
        \bottomrule
    \end{tabular}

\end{table}

These results suggest that diagnostic performance is relatively robust to ClinLab predictions, while the treatment of missing examinations can influence evaluation outcomes. Nevertheless, stable aggregate accuracy does not rule out individual cases where incorrect predicted values affect diagnostic reasoning. Further provenance-aware analysis is needed to distinguish decisions supported by observed EHR evidence from those influenced by ClinLab-generated results.

\section{Subgroup Robustness Analysis}
\label{app:subgroup}

We evaluate Qwen3-32B across patient-disjoint cohorts sampled from MIMIC-IV-Ext, covering different genders, age groups, and disease categories. Table~\ref{tab:subgroup_robustness} presents the subgroup results across five evaluation metrics.
As shown in Table~\ref{tab:subgroup_robustness}, diagnostic accuracy is relatively stable across age groups despite variations in information collection and examination coverage. Male and female cohorts also exhibit similar evidence-acquisition behavior but differ in diagnostic accuracy, indicating a disparity that warrants further investigation; unequal disease composition prevents a causal interpretation. More pronounced differences arise across diseases: heart failure is diagnosed substantially more accurately than sepsis, while pneumonia achieves the highest examination coverage without comparable diagnostic performance. These results suggest that \emph{evidence acquisition alone does not determine diagnostic success}; identifying and integrating clinically discriminative findings remain critical.

\begin{table}[t]
    \centering
    \caption{Subgroup performance of the fixed Qwen3-32B doctor agent. All values are percentages over completed evaluations. ICR, TOC, and ATC denote information collection rate, test-order coverage, and abnormal-test coverage, respectively. Category and subtype accuracy measure agreement at the ICD-10 block and three-character category levels. The urinary-system cohort includes N10--N16 and N30--N39.}
    \label{tab:subgroup_robustness}
    \small
    \resizebox{\columnwidth}{!}{
    \begin{tabular}{lrrrrr}
        \toprule
        Subgroup & ICR & TOC & ATC & Category Acc. & Subtype Acc. \\
        \midrule
        Male & 13.2 & 35.6 & 46.9 & 42.0 & 24.0 \\
        Female & 13.2 & 34.2 & 45.8 & 24.0 & 15.0 \\
        \midrule
        Age 18--44 & 15.2 & 39.2 & 48.5 & 33.0 & 18.0 \\
        Age 45--64 & 12.0 & 37.9 & 44.9 & 31.0 & 18.0 \\
        Age 65--74 & 12.9 & 38.6 & 45.7 & 30.0 & 18.0 \\
        Age 75+ & 11.1 & 41.0 & 51.2 & 33.0 & 16.0 \\
        \midrule
        Sepsis (A41) & 10.8 & 36.7 & 45.7 & 2.5 & 2.5 \\
        Heart failure (I50) & 14.4 & 40.3 & 45.2 & 71.0 & 71.0 \\
        Pneumonia (J18) & 12.9 & 51.5 & 62.9 & 28.0 & 27.0 \\
        Procedure complications (T81) & 12.2 & 31.8 & 41.0 & 23.0 & 15.0 \\
        Urinary-system diseases & 13.1 & 39.9 & 44.9 & 23.0 & 18.0 \\
        \bottomrule
    \end{tabular}
    }
\end{table}

We further examine three illustrative trajectories (Table~\ref{tab:subgroup_cases}). In the heart-failure case, targeted cardiac examinations reveal elevated NT-proBNP and pulmonary edema, supporting a correct diagnosis at the evaluated ICD-10 levels despite limited overall information collection. In contrast, the sepsis case focuses on biliary pathology, and the pneumonia case favors heart failure based on pulmonary edema and cardiomegaly; both disagree with the recorded primary diagnosis despite high examination coverage. Together, these examples illustrate that diagnostic performance depends not only on obtaining evidence but also on interpreting competing findings and aligning the final diagnosis with the underlying condition. However, the available observations do not establish that the reference diagnosis was unequivocally inferable in either mismatched case; examination results may also include laboratory-provider predictions.

\begin{table}[t]
    \centering
    \caption{Illustrative diagnostic trajectories across disease cohorts. All scores are percentages. Case identifiers refer to the frozen evaluation records. Reference and predicted diagnoses retain their original ICD-10 codes, while diagnostic accuracy is evaluated at the coarser levels defined in Section~\ref{sec:rule_based_scoring}.}
    \label{tab:subgroup_cases}
    \small
    \resizebox{\columnwidth}{!}{
    \begin{tabular}{llllrrrrr}
        \toprule
        Cohort & Case ID & Reference & Prediction & ICR & TOC & ATC & Category Acc. & Subtype Acc. \\
        \midrule
        Heart failure & \texttt{10199560\_22434262} & I50.23 & I50.1 & 6.4 & 43.1 & 100.0 & 100.0 & 100.0 \\
        Sepsis & \texttt{10651989\_24805766} & A41.9 & K80.3 & 19.5 & 67.4 & 100.0 & 0.0 & 0.0 \\
        Pneumonia & \texttt{10333909\_27924784} & J18.9 & I50 & 32.7 & 91.7 & 100.0 & 0.0 & 0.0 \\
        \bottomrule
    \end{tabular}
    }
\end{table}

\begin{table*}[t]
    \centering
    \caption{Representative successful and failed Qwen3-32B trajectories.}
    \label{tab:case_study}
    \small
    \setlength{\tabcolsep}{6pt}
    \renewcommand{\arraystretch}{1.12}
    \begin{tabular}{>{\raggedright\arraybackslash}p{0.14\textwidth}
                    >{\raggedright\arraybackslash}p{0.81\textwidth}}
        \toprule
        \rowcolor{green!6}
        \multicolumn{2}{l}{\textbf{Success: iterative evidence acquisition supports the correct ICD-10 category}} \\
        \midrule
        \textbf{Presentation} &
        A 57-year-old patient with hypertension, hyperlipidemia, diabetes,
        peripheral neuropathy, and current smoking presented with worsening
        positional dizziness, episodic severe hypertension, and bilateral hand
        paresthesia. Examination showed a left cervical bruit and subtle
        right-sided neurological abnormalities. \\
        \textbf{Diagnosis} &
        Agent output: \emph{cerebral infarction}, coded as I63. Reference:
        \emph{occlusion/stenosis of a carotid artery with cerebral infarction}
        (I63.239). \\
        \textbf{Metric scores} &
        $R_{\mathrm{info}}=$ 0.159 \quad
        $R_{\mathrm{test}}=$ 0.409 \quad
        $R_{\mathrm{abnormal}}=$ 0.111 \quad
        $A_{\mathrm{cat}}=$ 1 \quad
        $A_{\mathrm{sub}}=$ 1 \\
        \midrule
        \rowcolor{red!5}
        \multicolumn{2}{l}{\textbf{Failure: high evidence coverage still leads to a symptom-level answer}} \\
        \midrule
        \textbf{Presentation} &
        A 73-year-old patient with diabetes, diastolic heart failure,
        hypertension, possible coronary artery disease, and chronic kidney
        disease presented with progressive dyspnea and dry cough. Symptoms
        worsened with exertion and recumbency, but there was no chest pain. \\
        \textbf{Diagnosis} &
        Agent output: \emph{dyspnea and dry cough}, mapped to R06.0.
        Reference: \emph{subendocardial infarction/NSTEMI} (I21.4). \\
        \textbf{Metric scores} &
        $R_{\mathrm{info}}=$ 0.419 \quad
        $R_{\mathrm{test}}=$ 0.729 \quad
        $R_{\mathrm{abnormal}}=$ 0.667 \quad
        $A_{\mathrm{cat}}=$ 0 \quad
        $A_{\mathrm{sub}}=$ 0 \\
        \bottomrule
    \end{tabular}
\end{table*}

\begin{table*}[t]
    \centering
\caption{Turn-by-turn successful diagnostic trajectory for a de-identified case
(10006457\_27894366). D denotes Doctor, OPS Observable Patient Simulator,
LPS Latent Patient Simulator, LP Laboratory Physician, EC Environment
Controller, and DJ Diagnostic Judge. Patient responses are filtered by the
self-check mechanism before being returned to D, while clinical evidence is
provided through LP with support from LPS and \textsc{ClinLab} when needed.
Blue, orange, and green rows represent patient interaction, evidence
acquisition, and diagnosis submission, respectively.}
    \label{tab:case_interactions}
    \footnotesize
    \setlength{\tabcolsep}{5pt}
    \renewcommand{\arraystretch}{1.08}
    \begin{tabular}{>{\centering\arraybackslash}p{0.04\textwidth}
                    >{\raggedright\arraybackslash}p{0.16\textwidth}
                    >{\raggedright\arraybackslash}p{0.73\textwidth}}
        \toprule
        Turn & Information route & Actual interaction\\
        \midrule
        \rowcolor{black!5}
        Init. & EC $\rightarrow$ D &
        \textbf{EC:} Revealed a 57-year-old patient with hypertension,
        hyperlipidemia, diabetes with peripheral neuropathy, and current
        smoking. Admission examination included a left cervical bruit, subtle
        right-sided weakness and clumsiness, and an unsteady gait. \\
        \rowcolor{blue!4}
        1 & D $\leftrightarrow$ OPS &
        \textbf{D:} ``Describe your main symptoms, their onset and duration,
        and what worsens or relieves them.'' \quad
        \textbf{OPS:} ``I have recurrent dizziness, especially when standing.
        My blood pressure sometimes reaches 190--210, and both hands feel numb
        and tingly.'' \\
        \rowcolor{blue!4}
        2 & D $\leftrightarrow$ OPS &
        \textbf{D:} ``When did the dizziness begin, and was the onset sudden or
        gradual?'' \quad
        \textbf{OPS:} ``It began about one month ago, initially only
        occasionally, but has worsened over the past few days and is most
        noticeable whenever I stand.'' \\
        \rowcolor{blue!4}
        3 & D $\leftrightarrow$ OPS &
        \textbf{D:} ``Does the dizziness occur with nausea, blurred vision,
        palpitations, or sweating?'' \quad
        \textbf{OPS:} ``I sometimes have transient blurred vision, but no
        nausea, palpitations, or sweating.'' \\
        \rowcolor{orange!5}
        4 & \makecell[l]{D $\rightarrow$ LP\\LP $\leftrightarrow$ LPS\\LP $\rightarrow$ D} &
        \textbf{D:} Requested vital-sign and neurological examinations to
        evaluate dizziness, blood-pressure fluctuation, and paresthesia.
        \quad \textbf{LP (from LPS):} BP 115/74, left cervical bruit,
        distal sensory loss, slow/clumsy right-hand movements, and an unsteady
        gait. \\
        \rowcolor{orange!5}
        5 & \makecell[l]{D $\rightarrow$ LP\\LP $\leftrightarrow$ LPS/ClinLab\\LP $\rightarrow$ D} &
        \textbf{D:} Ordered a metabolic panel, cardiac markers, and
        inflammatory markers. \quad
        \textbf{LP (from LPS/ClinLab):} Glucose 214 mg/dL; creatinine
        0.8 mg/dL; sodium 138 mEq/L; potassium 4.3 mEq/L; troponin T
        $<0.01$ ng/mL; CK-MB 3.0 ng/mL. Additional requested items absent from
        the source record were generated by \textsc{ClinLab}. \\
        \rowcolor{orange!5}
        6 & \makecell[l]{D $\rightarrow$ LP\\LP $\leftrightarrow$ LPS\\LP $\rightarrow$ D} &
        \textbf{D:} Ordered non-contrast head CT to assess a possible
        cerebrovascular event. \quad
        \textbf{LP (from LPS):} No hemorrhage, edema, mass, or large-territory
        infarction; a small left frontal hypodensity could represent an old
        lacunar infarct or prominent perivascular space; intracranial vascular
        calcifications were present. \\
        \rowcolor{blue!4}
        7 & D $\leftrightarrow$ OPS &
        \textbf{D:} ``Have you noticed headache, speech difficulty, limb
        weakness, or other visual abnormalities?'' \quad
        \textbf{OPS:} ``I have a headache around my left eye, but no speech
        difficulty or limb weakness.'' \\
        \rowcolor{orange!5}
        8 & \makecell[l]{D $\rightarrow$ LP\\LP $\leftrightarrow$ LPS\\LP $\rightarrow$ D} &
        \textbf{D:} Ordered brain MRI to clarify the indeterminate left frontal
        CT finding. \quad
        \textbf{LP (from LPS):} Multiple diffusion-restricted foci consistent
        with infarction, including new left frontal and bilateral paramedian
        parietal lesions; a small amount of blood products was present in the
        left frontal focus. \\
        \rowcolor{green!5}
        Final & \makecell[l]{D $\rightarrow$ EC\\EC $\rightarrow$ DJ} &
        \textbf{D:} Submitted \emph{cerebral infarction} (I63). \quad
        \textbf{DJ:} The code matched the ICD-10 block and three-character
        category of the reference \emph{carotid occlusion/stenosis with
        cerebral infarction} (I63.239), although the agent did not specify the
        carotid mechanism. \\
        \bottomrule
    \end{tabular}
\end{table*}

\section{Case Study}
\textbf{Case Overview.}
We qualitatively analyze two Qwen3-32B trajectories from the same MIMIC-IV-Ext evaluation split used in the main experiments. To provide representative examples, we select cases with above-median information collection and test coverage among the completed trajectories. We include one successful case with a coherent evidence-acquisition process and one failed case with substantial evidence collection but an incorrect final diagnosis, allowing us to examine different sources of diagnostic errors. Table~\ref{tab:case_study} summarizes the selected cases. These examples are intended to illustrate the behaviors captured by the decomposed metrics rather than characterize their overall frequency.

We further presents the successful trajectory in a turn-by-turn format as shown in Table~\ref{tab:case_interactions}. Patient responses are translated into English, while lengthy laboratory and radiology reports are abbreviated with diagnostically relevant findings preserved. During interaction, the observable patient simulator provides patient-reportable information, whereas the latent patient simulator supplies privileged clinical states only through the laboratory physician. The doctor can therefore flexibly alternate between patient inquiry and objective evidence acquisition based on the evolving diagnostic hypothesis.

\textbf{Analysis.}
The successful case illustrates an evidence-driven but non-linear diagnostic process. After initial patient inquiry, the agent progressively requested neurological examination, laboratory tests, and imaging. When CT failed to provide definitive evidence, the agent conducted further targeted inquiry and ordered MRI, whose findings supported the final diagnosis at both category and subtype levels. However, it failed to identify the underlying vascular mechanism captured by the reference diagnosis, leading to incomplete abnormal-test coverage. This case demonstrates that \textsc{ClinVerse} can distinguish between achieving the required diagnostic granularity and fully characterizing the underlying pathology.

The failed case highlights a different limitation, where evidence acquisition was adequate but clinical interpretation was insufficient. Although the patient presented with risk factors and abnormal findings suggestive of ischemia, the agent failed to integrate these signals and terminated with only a symptom-level diagnosis. This example shows that collecting relevant evidence alone does not guarantee accurate diagnosis. By separately evaluating information acquisition and diagnostic reasoning, \textsc{ClinVerse} can identify whether failures arise from insufficient evidence gathering or ineffective evidence integration.

\end{document}